%% file: acl_2026.tex
\documentclass[11pt]{article}

\usepackage[preprint]{acl}

\usepackage{times}
\usepackage{latexsym}

\usepackage[T1]{fontenc}

\usepackage[utf8]{inputenc}

\usepackage{microtype}

\usepackage{inconsolata}

\usepackage{graphicx}

\input{math_commands.tex}

\usepackage{hyperref}
\usepackage{url}
\usepackage{newfloat}
\usepackage{listings}
\usepackage{amsfonts}       
\usepackage{nicefrac}       
\usepackage{microtype}      
\usepackage{amsmath}
\usepackage{subfig}
\usepackage{graphicx}
\usepackage{adjustbox}
\usepackage{booktabs}
\usepackage{multirow}
\usepackage{enumitem}
\usepackage{wrapfig}
\usepackage{natbib}
\newcommand{\our}{Short-RL}
\title{Shorten After You’re Right: Lazy Length Penalties for Reasoning RL}

\author{
\textbf{Danlong Yuan\textsuperscript{1,2}\thanks{Work done during internship at MSRA}},
 \textbf{Tian Xie\textsuperscript{3}\footnotemark[1]},
 \textbf{Shaohan Huang\textsuperscript{5}},
 \textbf{Zhuocheng Gong\textsuperscript{1}},
\\
 \textbf{Huishuai Zhang\textsuperscript{1,4}\footnotemark[2]},
 \textbf{ Chong Luo\textsuperscript{5}}
 \textbf{ Furu Wei\textsuperscript{5}}
\textbf{ Dongyan Zhao\textsuperscript{1,4}\footnotemark[2]}
\\
 \textsuperscript{1}Wangxuan Institute of Computer Technology, Peking University,
 \\
 \textsuperscript{2}Center for Data Science, AAIS, Peking University,
 \\
 \textsuperscript{3}University of Science and Technology of China,
\\
 \textsuperscript{4}State Key Laboratory of General Artificial Intelligence,
 \\
 \textsuperscript{5}Microsoft Research
 \\
  \small{
  \textbf{Code:} \href{https://github.com/lblankl/Short-RL}{https://github.com/lblankl/Short-RL}
  }
}

\begin{document}
\maketitle
\begin{abstract}
Long-reasoning models achieve strong accuracy on complex reasoning tasks, but their extended reasoning trajectories incur substantial memory and latency costs. Several existing shortening methods rely on additional supervision or multi-stage post-training, which primarily reduces inference length and does not reduce the rollout tokens during on-policy reinforcement learning (RL). We instead target \emph{on-policy} response shortening, aiming to improve both inference efficiency and RL training throughput. However, because on-policy RL couples optimization with exploration, naively penalizing length can destabilize training and suppress exploration. To impose length pressure safely, we propose a \emph{lazy length penalty} integrated into the rule-based RL pipeline: it activates only on correct trajectories, only after training accuracy enters a stably improving regime, and only when responses exceed a \emph{tolerance band} beyond the minimal correct length. Across four settings, our method significantly reduces response length without extra training stages while maintaining or improving performance. In a logic reasoning setting, we achieve a 40\% reduction in step-averaged response length alongside a 14-point gain in performance. For math problems, we reduce step-averaged response length by 33\% while preserving performance.
\end{abstract}


\section{Introduction}

Long reasoning models (LRMs) trained with large-scale, rule-based on-policy reinforcement learning (RL) have achieved strong performance on complex reasoning tasks, often exhibiting self-reflection and self-correction~\cite{longsurvey,deepseekai2025deepseekr1incentivizingreasoningcapability,openai2024openaio1card}. A recurring empirical trend in these pipelines is that reasoning trajectories grow longer as training progresses, which can correlate with improved accuracy~\cite{deepseekai2025deepseekr1incentivizingreasoningcapability}. However, this length growth is costly: longer outputs increase inference latency and KV-cache memory, and longer rollouts directly reduce RL training throughput, sometimes making large-scale on-policy RL impractical.

Importantly, longer reasoning is not always better. Models can overthink, repeat steps, or drift into redundant verification, and unnecessary length may even hurt performance~\cite{overthinking25,kimi2025,yang2025dynamicearlyexitreasoning,sui2025stopoverthinkingsurveyefficient}. These observations motivate methods that shorten reasoning trajectories while preserving correctness.

Most existing shortening approaches rely on extra supervision, distillation, or off-policy/post-training stages~\cite{tokenskip2025,C3ot2025,cotValue2025,selftrain2025,distilling2to12024,skipsteps2024,stepre2025,o1-pruner2025,DAST2025}. Such methods can reduce inference length, but they do not reduce the rollout tokens already spent during the main on-policy RL stage. A natural alternative is to directly penalize length in the on-policy reward (e.g., Kimi~\cite{kimi2025}). Yet in our reproduction, applying length shaping early causes a failure mode: trajectories collapse to overly short outputs, exploration is suppressed, and training becomes unstable, leading to suboptimal performance. Similar degradation is reported elsewhere~\cite{Arora2025TrainingLM,ThinkPrune}. This suggests that on-policy length control must respect the coupling between exploration and optimization.

We take the view that \textbf{length is an auxiliary property of a reasoning trajectory}: success is defined by correctness, while brevity is a preference \emph{among successful trajectories}. Hence, length regularization in on-policy RL should be \emph{lazy}: apply it \textbf{where} it is safe (only on correct trajectories), \textbf{when} learning is stable, and \textbf{what} is truly redundant (only excess length beyond a tolerance band).

Guided by this principle, we propose \our{}, a lazy length penalty integrated into rule-based on-policy RL. It consists of three gates: \textsc{RightGate} applies shaping only to correct trajectories, \textsc{SlackBand} penalizes only excess length beyond a tolerated minimum, and \textsc{StableSwitch} activates shaping only once batch accuracy is sufficiently stable. Across four settings, \our{} significantly shortens trajectories without extra training stages while maintaining or improving performance. On Logic-RL, we reduce step-averaged response length by 40\% while improving performance by 14 points; on three math RL pipelines, we reduce step-averaged length by up to 33\% while preserving performance.

\section{Related Work}

We build on (i) long-reasoning models trained with large-scale, rule-based on-policy RL, and (ii) methods for shortening long-form reasoning. Our focus is the intersection: \emph{shortening trajectories during on-policy RL} without extra post-training stages.

\subsection{Rule-Based RL for Long Reasoning}
Recent LRMs improve performance by generating explicit multi-step trajectories and are often trained with large-scale, rule-based on-policy RL, which can enhance reasoning without human preference labels~\cite{deepseekai2025deepseekr1incentivizingreasoningcapability,openai2024openaio1card}. A consistent empirical observation is that response length tends to increase during training, raising inference latency and KV-cache usage, and more critically, increasing rollout token cost during RL~\cite{deepseekai2025deepseekr1incentivizingreasoningcapability}. This trend appears across domains, including logic reasoning (e.g., Logic-RL)~\cite{xie2025logicrlunleashingllmreasoning,Xie2024OnMO} and mathematical reasoning pipelines~\cite{simplerlmath2025,deepscaler2025,OpenReasonerZero2025,dapo2025}, motivating efficiency improvements that operate \emph{within} the on-policy training loop.

\subsection{Shortening Long-Form Reasoning}
A broad literature seeks to reduce the cost of long-form reasoning by shortening trajectories or reducing generated tokens.

\paragraph{Supervised and off-policy shortening.}
Many methods rely on supervised fine-tuning or distillation to compress chains-of-thought or skip steps~\cite{tokenskip2025,C3ot2025,cotValue2025,selftrain2025,distilling2to12024,skipsteps2024,stepre2025}, or use off-policy RL / extra post-training stages to prune and revise traces~\cite{o1-pruner2025,DAST2025}. While effective, these approaches typically require additional data or training phases and are not directly aligned with the \emph{in-process} on-policy RL pipelines used to train LRMs.

\paragraph{Inference-time control.}
Prompting and decoding-time strategies (e.g., concise prompting, token budgets, early stopping) reduce inference cost by limiting or guiding generation~\cite{2025tokenbudget,2024Concise,2025chaindraft,NoThink}. Related work on routing and dynamic computation allocates reasoning effort based on difficulty or intent~\cite{Claude3.7,Sketch-of-Thought,confidence,RouteLLMs,THOUGHTTERMINATOR,L1,modelcollaboration,modelmerging}. These methods primarily affect inference and do not reduce the rollout tokens already consumed during RL training.

\paragraph{On-policy length rewards.}
Closest to our setting are length-based rewards integrated into RL. Kimi~\cite{kimi2025} proposes a direct length reward but applies it in a \emph{post-RL} stage, noting that using it early can harm training. Efficient~\cite{Arora2025TrainingLM} scales rewards based on response length and observes accuracy--length trade-offs, while ThinkPrune~\cite{ThinkPrune} penalizes correct responses beyond a limit and similarly highlights the brevity--accuracy tension.

\paragraph{Our position.}
In contrast to post-RL or inference-only approaches, we target \emph{training-time} efficiency in on-policy RL. We treat length as an auxiliary trajectory property and introduce a \emph{lazy} length penalty that (i) applies only on correct trajectories, (ii) penalizes only excess length beyond a tolerance band, and (iii) activates only after training becomes stable. This design aims to shorten rollouts \emph{during} RL training while avoiding the exploration and stability failures of naive always-on length shaping.

\section{Methodology}

\subsection{Notation and Setup}
Let \(x\) be the prompt and \(y^*\) the reference answer. Each rollout produces a full output \(\hat{u}_i\), parsed into a reasoning trace \(z_i\) and final answer \(y_i\). We define correctness \(c_i := \mathbb{I}[y_i=y^*]\) and length \(l_i := l(\hat{u}_i)\) (token count). Let \(\text{acc}\) be the batch correctness rate over rollout samples and \(\text{acc}_{\max}\) its running maximum.

\subsection{Background: Length-Aware Rewards in Rule-Based On-Policy RL}
\begin{figure*}[htbp]
  \centering
  \subfloat[Kimi]
  {
      \includegraphics[width=0.3\linewidth]{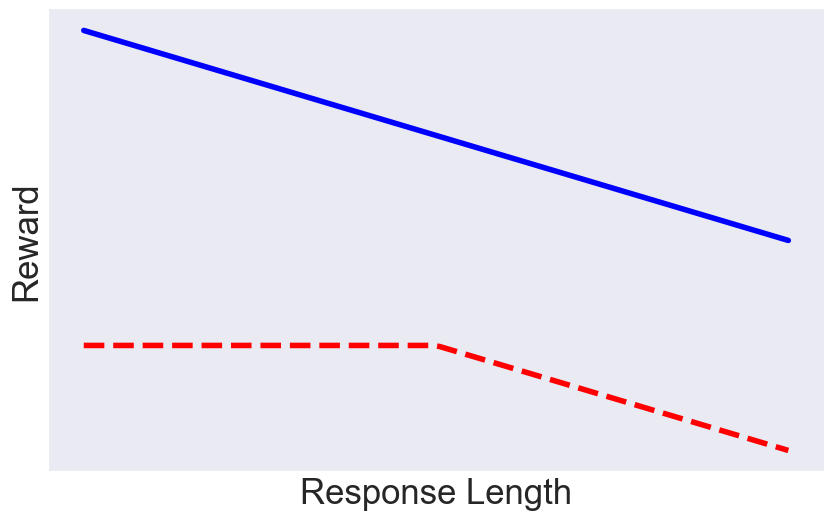}
      \label{figcmp:subfig1}
  }
  \subfloat[Efficient]
  {
      \includegraphics[width=0.3\linewidth]{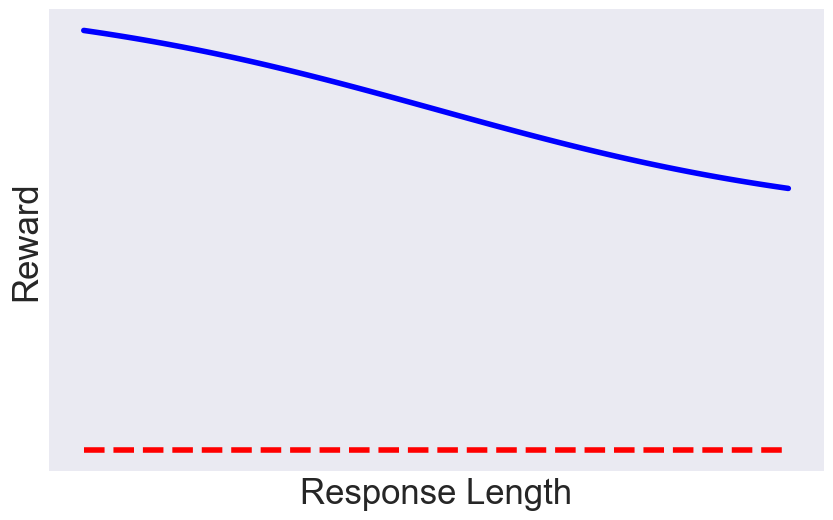}
       \label{figcmp:subfig2}
  }
  \subfloat[ThinkPrune]
  {
      \includegraphics[width=0.3\linewidth]{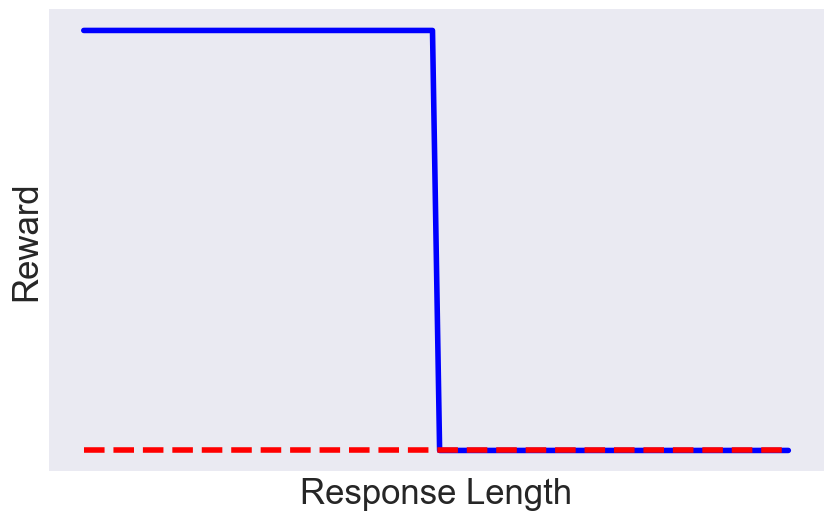}
       \label{figcmp:subfig3}
  }
  \caption{Reward values as a function of response length. Blue lines indicate rewards for correct responses and red lines represent rewards for incorrect responses.}
  \label{figcmp}
\end{figure*}

A standard way to encourage shorter trajectories is to add a length term to the original task reward:
\begin{equation}
\label{eq:base_reward}
R(x,\hat{u}) = R_{\text{task}}(x,\hat{u}) + \alpha \cdot R_{\text{len}}(x,\hat{u}),
\end{equation}
where \(R_{\text{task}}\) is the original rule-based reward (e.g., correctness/format), \(R_{\text{len}}\) is a shaping term based on output length, and \(\alpha\) controls its strength. Figure~\ref{figcmp} illustrates representative designs (Kimi, Efficient, ThinkPrune).

Among them, Kimi~\cite{kimi2025} is a common choice and serves as our case study. For each prompt \(x\), sample \(k\) outputs and compute \(l_{\min}=\min_i l_i\), \(l_{\max}=\max_i l_i\). If \(l_{\max}=l_{\min}\), set the length reward to \(0\). Otherwise,
\begin{equation}
\label{eq:kimi_len}
R_{\text{len}}(i)=
\begin{cases}
\lambda_i, & c_i=1,\\
\min(0,\lambda_i), & c_i=0,
\end{cases}
\end{equation}
where $\lambda_i = 0.5 - \frac{l_i-l_{\min}}{l_{\max}-l_{\min}}.
$

\begin{figure}[!h]
  \centering
  \subfloat
  {
      \includegraphics[width=0.7\linewidth]{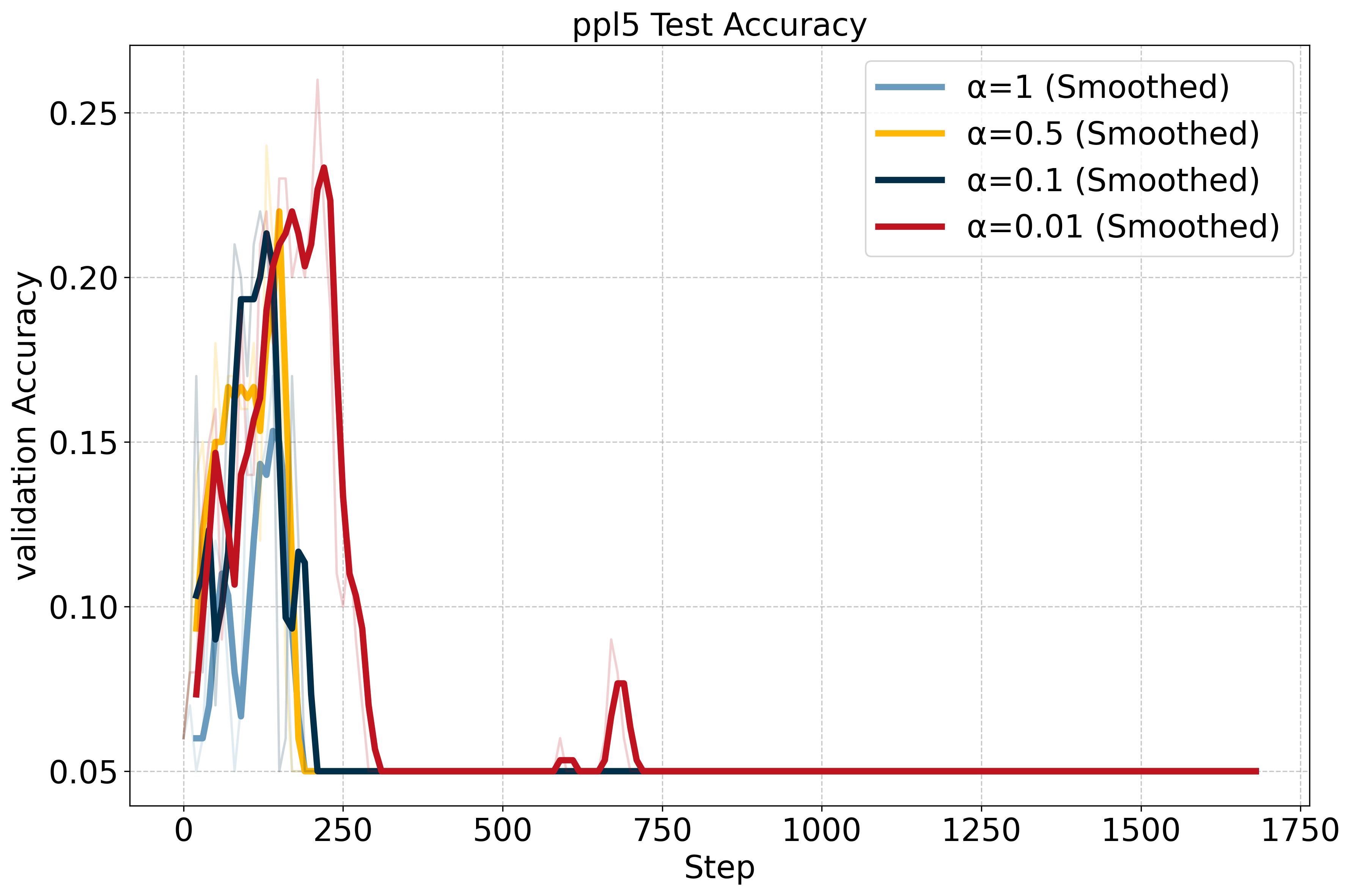}
      \label{figkimi:subfig1}
  }\\
  \subfloat
  {
      \includegraphics[width=0.7\linewidth]{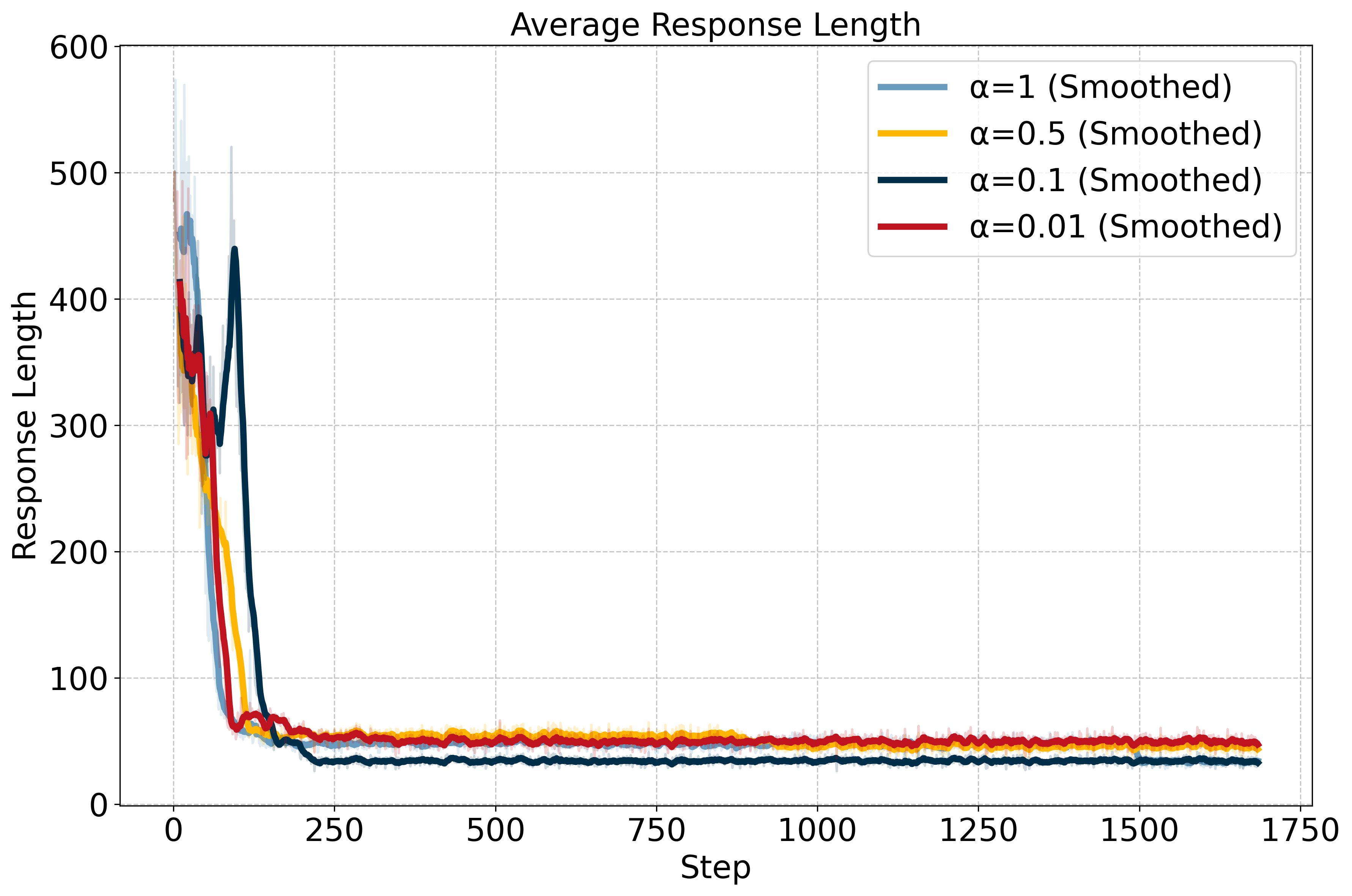}
       \label{figkimi:subfig2}
  }
  \caption{Test accuracy (left) and average response length (right) across different \(\alpha\).}
  \label{figkimi}
\end{figure}

\subsection{Why Naive On-Policy Length Shaping Fails}
\label{sec:failures}
Length is an \emph{auxiliary} trajectory attribute: correctness defines success, while shorter outputs are preferred \emph{among correct trajectories}. In on-policy RL, optimization and data collection are coupled, so always-on length shaping can reshape the sampled trajectory distribution and interfere with exploration.

\paragraph{Empirical early collapse.}
Using Kimi-style shaping from the start of Logic-RL training~\cite{xie2025logicrlunleashingllmreasoning} causes reward-hacking: trajectories rapidly collapse to very short outputs and accuracy becomes unstable or degrades (Figure~\ref{figkimi}).

\begin{figure}[h]
  \centering
  \includegraphics[width=0.8\linewidth]{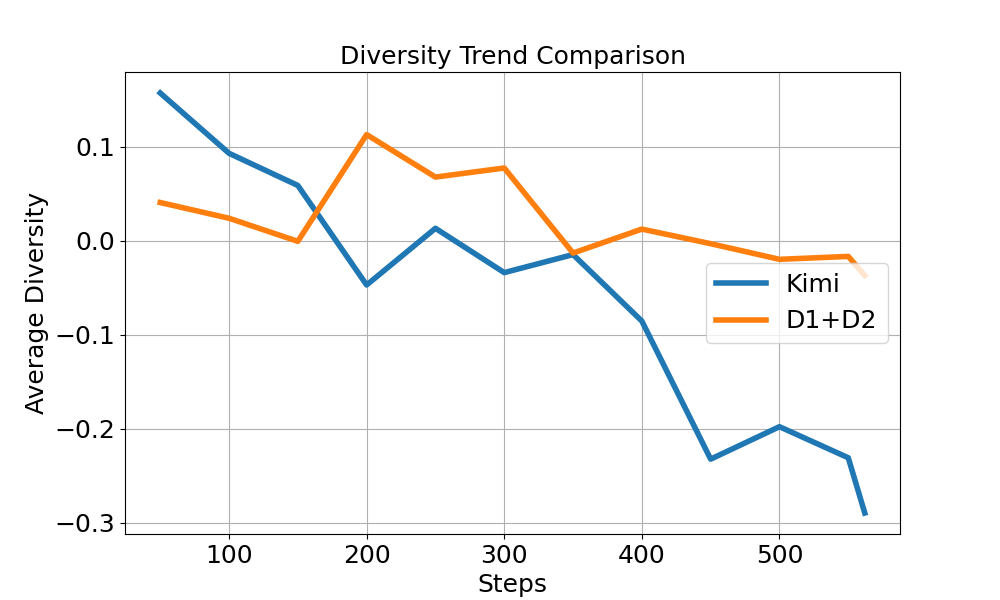}
  \caption{Diversity metric on \texttt{ppl5} during training.}
  \label{fig:diversity}
\end{figure}

\paragraph{Two conflicts.}
(i) \textbf{Exploration suppression:} Eq.~\ref{eq:kimi_len} computes \(l_{\min},l_{\max}\) over \emph{all} samples and can further penalize incorrect rollouts, biasing sampling toward short (often uninformative) trajectories and reducing reasoning-path diversity (Figure~\ref{fig:diversity}).  
(ii) \textbf{Training instability:} applying length pressure throughout training can compete with competence acquisition, since longer traces may be necessary to discover correct strategies at certain stages.

These observations motivate a simple principle: \emph{auxiliary-trajectory regularization should be lazy}—it should specify \textbf{where} it applies (only correct trajectories), \textbf{what} it penalizes (only excess length), and \textbf{when} it activates (only after learning is stable).

\subsection{\our{}: Lazy Length Penalties}
We implement this principle via three gates:
\begin{itemize}[leftmargin=15pt]
\item \textbf{\textsc{RightGate} (Where).} Apply length shaping only to correct trajectories. For each prompt \(x\), compute \(l_{\min}\) and \(l_{\max}\) using only correct samples \(\mathcal{C}(x)=\{j\mid c_j=1\}\), and set the length term to \(0\) for incorrect samples.
\item \textbf{\textsc{SlackBand} (What).} Penalize only \emph{excess} length beyond a tolerance \(\tau_l\). If \(l_i \le l_{\min}+\tau_l\), we apply a constant baseline (no preference among these correct trajectories); only samples exceeding the band receive decreasing reward.
\item \textbf{\textsc{StableSwitch} (When).} Activate length shaping only when training is stable: enable the length term only if \(\text{acc} \ge \text{acc}_{\max} - \tau_{\text{acc}}\).
\end{itemize}

\subsection{Unified Reward}
Putting the gates together, the length shaping term for sample \(i\) is
\begin{equation}
\label{eq:lazy_len}
\begin{aligned}
R_{\text{len}}(i) &=
\begin{cases}
\beta_i, & \text{if } c_i = 1\text{,} \;\; \text{acc} \ge \text{acc}_{\max} - \tau_{\text{acc}},\\
0, & \text{otherwise},
\end{cases}\\
\beta_i &=
\begin{cases}
\lambda_i, & \text{if } l_i > l_{\min} + \tau_{l},\\
0.5, & \text{otherwise},
\end{cases}\\
\lambda_i &= 0.5 - \frac{l_i-l_{\min}}{l_{\max}-l_{\min}},
\end{aligned}
\end{equation}
with $\mathcal{C}(x) = \{j \mid c_j = 1\}$ and 
\begin{equation*}
l_{\min} = \min_{j \in \mathcal{C}(x)} l_j, \qquad
l_{\max} = \max_{j \in \mathcal{C}(x)} l_j.
\end{equation*}
If \(|\mathcal{C}(x)|=0\), we set \(R_{\text{len}}(i)=0\) for all \(i\). If \(l_{\max}=l_{\min}\), the ``excess-length'' condition is never satisfied, so \(\lambda_i\) is unused and correct samples default to the constant baseline.

Finally, we plug \(R_{\text{len}}\) into Eq.~\ref{eq:base_reward}. The tolerance \(\tau_l\) controls how much length is ignored as acceptable slack, while \(\tau_{\text{acc}}\) controls how early and how frequently length shaping activates.

\section{Experiments}
In this section, we evaluate whether \our{} can \emph{shorten reasoning trajectories during on-policy RL} while maintaining (or improving) task performance. Since the dominant cost of on-policy RL scales with rollout tokens, we emphasize both \emph{accuracy} and \emph{efficiency}. We report two complementary length metrics: \textbf{Training (step-avg)} as a proxy for training-time token cost of a response, and \textbf{Inference (final)} as a proxy for test-time decoding cost for one question.

\subsection{Experimental Settings}
We first describe the evaluation domains, training setups, metrics, and baselines used throughout the experiments.

\paragraph{Evaluation domains and training setups.}
We evaluate on two domains: logic reasoning and mathematical reasoning. The logic domain follows Logic-RL~\citep{xie2025logicrlunleashingllmreasoning}. The math domain includes three representative on-policy RL pipelines: DeepScaleR~\citep{deepscaler2025}, SimpleRL-Reason~\citep{simplerlmath2025}, and Open-Reasoner-Zero~\citep{OpenReasonerZero2025}. Across all experiments, we use the same model architecture and training framework as the corresponding original projects in the VeRL framework~\citep{verl}. For the three math settings, we adopt a DeepSeek-R1-style prompt template and include the format reward used by the original pipelines. Full prompts and hyperparameters are in Appendix~\nameref{details}.

\paragraph{Training-cost and inference-cost metrics.}
On-policy RL cost scales with the number of generated tokens in rollouts and overlong responses require significant memory cost and computation time. To make efficiency explicit, we report two length metrics:
\textbf{(i) Training (step-avg)}: response length averaged over training steps (proxy for training-time token cost and throughput);
\textbf{(ii) Inference (final)}: response length measured at the final checkpoint (proxy for test-time decoding cost).
Unless stated otherwise, accuracy is always evaluated at the final checkpoint.

\subsubsection{Logic Reasoning}
We begin with logic reasoning, where long rollouts naturally emerge and the training-time efficiency benefit of on-policy shortening is most visible.
We use the same datasets and evaluation protocol as Logic-RL~\citep{xie2025logicrlunleashingllmreasoning}, initializing from Qwen2.5-7B~\citep{qwen2025}. For \our{}, we set $\tau_{l}=200$, $\tau_{\text{acc}}=0.05$, and $\alpha=1$ (details in Appendix~\nameref{details}). We report in-domain accuracy on 2- to 8-person tasks, and out-of-domain generalization on AMC and AIME following the Logic-RL protocol. We also report Training (step-avg) and Inference (final) lengths.

\subsubsection{Math Reasoning}
We next evaluate on mathematical reasoning, which provides three different RL pipelines and tests whether the same lazy length penalty generalizes across implementations.
We evaluate on DeepScaleR~\citep{deepscaler2025}, Open-Reasoner-Zero~\citep{OpenReasonerZero2025}, and SimpleRL-Reason~\citep{simplerlmath2025}, reusing nearly all hyperparameters from each original implementation. \our{} hyperparameters for each setting are listed in Appendix Table~\ref{tab:detail}. We evaluate on five benchmarks: AIME2024~\citep{AIME2024}, AMC23~\citep{AMC23}, MATH-500~\citep{math500}, Minerva Math~\citep{math}, and Olympiad Bench~\citep{OlympiadBench}. We again report Training (step-avg) and Inference (final) lengths.

\subsubsection{Baselines}
Finally, we describe the baselines used to isolate the effects of on-policy length shaping and to compare against prior length-control rewards.
We compare against:
\begin{itemize}[leftmargin=15pt]
\item \textbf{Standard}: on-policy RL with the original task reward \(R_{\text{task}}\) (no length shaping).
\item \textbf{Kimi (post)}: the two-stage procedure used by Kimi~\citep{kimi2025}: first run Standard RL, then apply the length reward in a later stage. \textbf{Importantly}, this post-RL setup can shorten \emph{inference} trajectories, but it does \emph{not} reduce the rollout tokens already consumed in stage-1 RL. Therefore, in our tables we report the \textbf{Training (step-avg)} length of the first (Standard) stage for Kimi (post), as this reflects the actual training-time cost.
\item \textbf{Efficient}: length-aware scaling reward from~\citep{Arora2025TrainingLM}. We tune the scaling coefficient over $\{0.02,0.05,0.08,0.10\}$ per setting: Logic-RL ($0.05$), DeepScaleR ($0.10$), and Open-Reasoner-Zero / SimpleRL-Reason ($0.02$). Note this coefficient differs from our \(\alpha\).
\item \textbf{ThinkPrune}: cosine length reward from~\citep{ThinkPrune}. We select a length limit that yields a comparable \textbf{Inference (final)} length to \our{}: 1700 (Logic-RL), 2500 (DeepScaleR), and 1500 (Open-Reasoner-Zero / SimpleRL-Reason).
\end{itemize}

\subsection{Main Results}
We now present the main quantitative results. We start with logic reasoning, highlighting both accuracy and training-time token savings, and then move to math reasoning to evaluate generality across three pipelines.

\subsubsection{Logic Reasoning: Accuracy Improves While Cost Drops}
Table~\ref{tab:logicrl} shows that \our{} consistently improves accuracy while significantly reducing training-time rollout tokens. Compared with Standard RL, \our{} reduces \textbf{Training (step-avg)} length by 40\% (1477$\rightarrow$889) while improving average in-domain accuracy by 14 points (79$\rightarrow$93). Inference efficiency also improves substantially (2632$\rightarrow$535).

A key distinction from post-RL length control is training efficiency: Kimi (post) reduces \textbf{Inference (final)} length, but it inherits the \textbf{Training (step-avg)} length of the initial Standard RL stage, and therefore does not reduce training-time rollout cost. In contrast, \our{} applies a lazy length penalty \emph{on-policy} after the policy reaches stable competence, shortening trajectories during RL itself and directly reducing training cost---analogous in spirit to trajectory truncation in standard RL, but achieved through a learned, stability-aware mechanism rather than a hard cap.

In addition, Figure~\ref{fig:len_control} shows that our penalty is activated \emph{on-policy} during RL: once accuracy becomes stable, the length control rate rises and the average rollout length decreases, explaining the reduction in Training (step-avg) tokens without relying on a separate post-RL stage.

\begin{table*}[t]
  \setlength{\tabcolsep}{3.0pt}

  \centering
  \begin{adjustbox}{width=2 \columnwidth,center}
  \begin{tabular}{lcccccccccccc}
  \toprule
  \multirow{2}{*}{\textbf{Method}}  & \multicolumn{8}{c}{\textbf{In Domain}}  & \multicolumn{2}{c}{\textbf{Out of Domain}} & \multicolumn{2}{c}{\textbf{Avg. Response Length}} \\
  \cmidrule(r){2-9} \cmidrule(r){10-11} \cmidrule(r){12-13}
  & \bf ppl2 & \bf ppl3 & \bf ppl4 & \bf ppl5 & \bf ppl6 & \bf ppl7 & \bf ppl8 & \bf Avg
  & \bf AMC & \bf AIME & \bf Training (step-avg) & \bf Inference (final) \\
\midrule
 Standard    & 82 &87 &88 &81 &76 &69 &70 &79 &39.76 &7.77  &1477 &2632   \\
 Kimi (post) & 84 &88 &89 &84 &79 &74 &76 &82 &39.89 &8.13  &1477 &763   \\
 Efficient   & 76 &81 &79 &77 &62 &48 &51 &68 &37.35 &7.77  &772  &843   \\
 ThinkPrune  & 80 &84 &86 &82 &70 &66 &64 &76 &38.47 &7.35  &832  &793   \\
 \our{}      & \textbf{97} &\textbf{97} &\textbf{99} &\textbf{95} &\textbf{92} &\textbf{83} &\textbf{87} &\textbf{93} &\textbf{42.17} &\textbf{8.74}  &889 &535 \\
  \bottomrule
  \end{tabular}
  \end{adjustbox}
  \caption{Logic-RL evaluation at the final checkpoint. ``Training (step-avg)'' is the response length averaged over training steps (proxy for training-time token cost); ``Inference (final)'' is the response length at the final checkpoint (proxy for test-time decoding cost).}
\label{tab:logicrl}
\end{table*}

\subsubsection{Math Reasoning: Cost Savings Without Sacrificing Accuracy}
We next turn to math reasoning to test robustness across three different RL pipelines and evaluation suites.
Table~\ref{tab:allmath} shows that \our{} yields consistent reductions in \textbf{Training (step-avg)} length while preserving (and occasionally improving) average accuracy. Relative to Standard RL, \our{} reduces Training (step-avg) length by 33\%, 11\%, and 21\% on DeepScaleR, Open-Reasoner-Zero, and SimpleRL-Reason, respectively. Efficient and ThinkPrune typically shorten outputs more aggressively but exhibit clearer accuracy--length trade-offs. Similar to logic reasoning, Kimi (post) can reduce \textbf{Inference (final)} length but does not reduce training-time rollout cost because its first-stage RL is unchanged.

We further inspect training dynamics in Figure~\ref{fig:len_control}: after competence stabilizes, the length control rate increases and the rollout length decreases, confirming that the savings in Training (step-avg) length come from \emph{on-policy} shortening rather than post-hoc compression.

\begin{table*}[t]
  \setlength{\tabcolsep}{2.0pt}
  \centering
  \begin{adjustbox}{width=2.0 \columnwidth,center}
  \begin{tabular}{lcccccccc}
  \toprule
  \multirow{2}{*}{\textbf{Method}}  & \multicolumn{6}{c}{\textbf{Math Benchmarks}}   & \multicolumn{2}{c}{\textbf{Avg. Response Length}} \\
  \cmidrule(r){2-7} \cmidrule(r){8-9}
  & \bf AIME2024 &\bf AMC23 &\bf MATH\-500 &\bf Minerva Math &\bf Olympiad Bench & \bf Avg
  & \bf Training (step-avg) & \bf Inference (final) \\
\midrule
\multicolumn{9}{l}{\textit{DeepScaleR}} \\
\midrule
 Standard    & 26.67 &59.04 &\pmb{81.40} &\pmb{26.10} &42.65 &47.17 &2523 &3072 \\
 Kimi (post) & 23.33 &\pmb{61.45} &81.00 &25.37 &\pmb{42.79} &46.79 &2523 &1678 \\
 Efficient   & 20.00 &49.40 &57.8 &16.54 &33.73 &35.49 &1517 &1537 \\
 ThinkPrune  & 26.67 &56.63 &78.40 &25.74 &41.31 &45.75 &1589 &1621 \\
 \our{}      & \pmb{30.00} & \pmb{60.24} &80.60 &\pmb{26.47} &\pmb{42.65} &\pmb{47.99} &1692 &1700 \\
\midrule
\multicolumn{9}{l}{\textit{Open-Reasoner-Zero}} \\
\midrule
 Standard    & 16.67 &\pmb{50.60} &\pmb{78.80} &30.88 &38.04 &43.00 &746 &840 \\
 Kimi (post) & \pmb{20.00} &49.40 &77.40 &\pmb{31.25} &\pmb{38.63} &\pmb{43.34} &746 &621 \\
 Efficient   & 13.33 &46.99 &66.40 &26.47 &35.96 &37.83 &578 &655 \\
 ThinkPrune  & 13.33 &48.19 &76.80 &27.57 &37.15 &40.61 &677 &682 \\
 \our{}      & 16.67 &\pmb{50.60} &78.60 &30.52 &38.19 &42.92 &660 &670 \\
\midrule
\multicolumn{9}{l}{\textit{SimpleRL-Reason}} \\
\midrule
 Standard    & 13.33 &48.19 &77.00 &\pmb{32.72} &\pmb{39.97} &42.24 &703 &791 \\
 Kimi (post) & 16.67 &48.19 &77.40 &31.99 &39.67 &42.78 &703 &601 \\
 Efficient   & 6.67  &38.55 &64.8  &22.06 &28.68 &32.15 &492 &532 \\
 ThinkPrune  & 10.00 &46.99 &69.40 &31.62 &37.30 &39.06 &613 &598 \\
 \our{}      & \pmb{20.00} &\pmb{49.40} &\pmb{78.20} &\pmb{32.72} &39.23 &\pmb{43.91} &554 &620 \\
\bottomrule
  \end{tabular}
  \end{adjustbox}
  \caption{Math reasoning evaluation. ``Training (step-avg)'' proxies training-time rollout token cost; ``Inference (final)'' is the final-checkpoint response length.}
\label{tab:allmath}
\end{table*}

\subsection{Training Dynamics: When Does the Lazy Length Penalty Activate?}
\label{track}
The main results above emphasize that \our{} reduces \textbf{Training (step-avg)} length---and therefore training-time rollout token cost---which is fundamentally different from post-RL shortening (e.g., Kimi (post)) that cannot reduce tokens already spent in the initial RL stage. We now make this difference explicit by analyzing \emph{when} the lazy length penalty is activated during training. This analysis directly reflects the behavior of \textsc{StableSwitch} (stability-gated activation) and \textsc{SlackBand} (penalize only excess length): early training should prioritize exploration and competence acquisition, while shortening should take effect only after the policy becomes reliably correct.

\paragraph{Length control rate.}
To quantify how frequently length shaping is applied, we introduce a batch-wise metric called \emph{length control rate} \(\gamma_{l}\).
For each batch, let \(N\) be the number of \emph{correct} responses (i.e., number of samples with \(c_i=1\)). Among them, let \(R\) be the number of correct responses that receive a strict length penalty, i.e., those with \(R_{\text{len}}(i) < 0.5\). Since \(R_{\text{len}}(i)<0.5\) occurs exactly when a correct trajectory exceeds the tolerance band (\textsc{SlackBand}), \(\gamma_{l}\) measures the fraction of correct trajectories that are actively shortened. We define:
\begin{equation}
\gamma_{l} =
\begin{cases}
-1, & \text{if } \text{acc} < \text{acc}_{\max} - \tau_{\text{acc}}, \\
0, & \text{if } N = 0, \\
\frac{R}{N}, & \text{otherwise}.
\end{cases}
\end{equation}
Here, \(\gamma_{l}=-1\) indicates that \textsc{StableSwitch} disables the length penalty due to unstable accuracy; \(\gamma_{l}\in[0,1]\) indicates that the penalty is enabled and quantifies its active rate on correct rollouts.

\begin{figure}[htbp]
  \centering
  \subfloat[Logic-RL]
  {\includegraphics[width=0.75\linewidth]{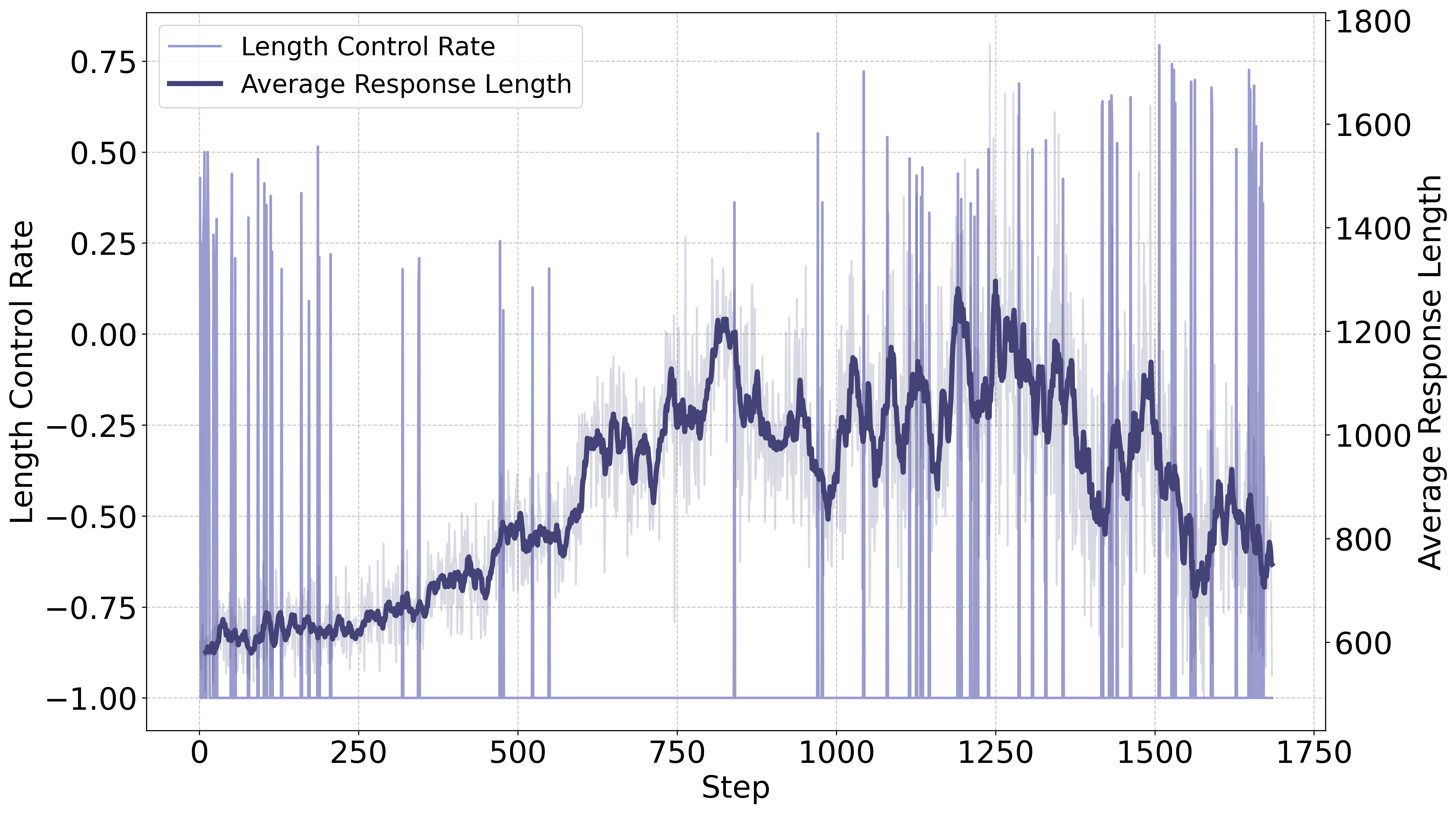}}\\
  \subfloat[DeepScaleR]
  {\includegraphics[width=0.75\linewidth]{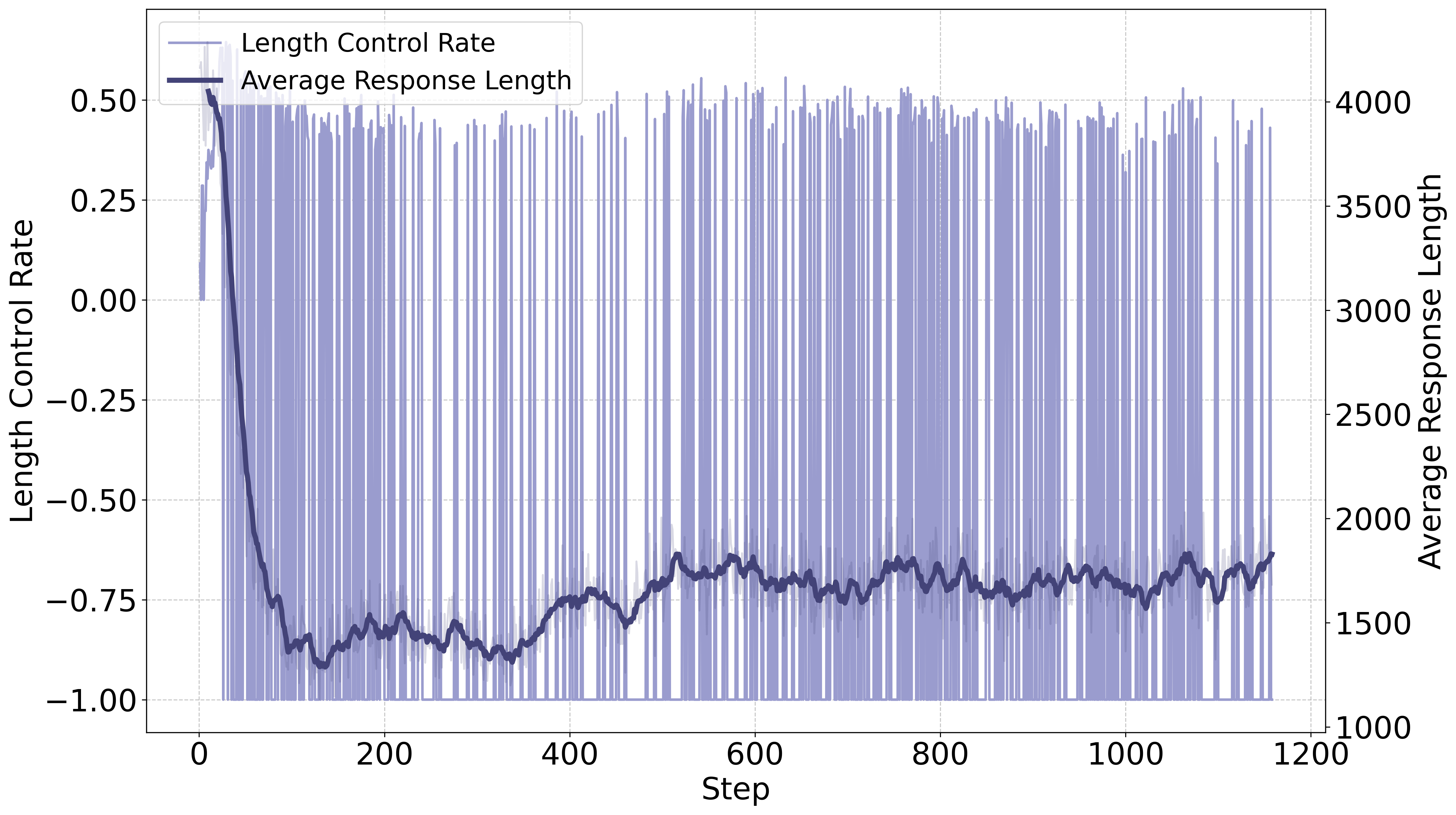}}
  \caption{Tracking the lazy length penalty during training. We plot the length control rate \(\gamma_{l}\) and the average response length over training steps. \(\gamma_{l}=-1\) indicates that \textsc{StableSwitch} disables length shaping; \(\gamma_{l}\in[0,1]\) measures the fraction of correct trajectories that are penalized for exceeding the tolerance band.}
  \label{fig:len_control}
\end{figure}

\paragraph{From stable accuracy to shorter rollouts.}
Figure~\ref{fig:len_control} tracks \(\gamma_{l}\) and the average response length during training. We observe a consistent pattern: in early training, \(\gamma_{l}\) is frequently \(-1\), meaning that the length penalty is intentionally \emph{inactive} while the model is still learning to solve the task. As training progresses and batch accuracy stabilizes, \textsc{StableSwitch} enables length shaping (\(\gamma_{l}\ge 0\)), and a non-trivial fraction of correct trajectories are penalized for exceeding the tolerance band (\(\gamma_{l}>0\)). This coincides with a clear reduction in average rollout length, explaining why \our{} reduces \textbf{Training (step-avg)} length and thus training cost.

\paragraph{Why on-policy matters.}
This behavior contrasts with post-RL shortening: because \our{} applies the penalty \emph{during} the main RL stage (but only after the policy is ``right'' and stable), it reduces rollout tokens for subsequent updates, aligning with the practical role of truncating overlong trajectories in standard RL. In DeepScaleR, we observe a higher length control rate, indicating that more correct trajectories exceed the tolerance band once competence stabilizes, which leads to stronger training-time savings. Curves for SimpleRL-Reason and Open-Reasoner-Zero are provided in Appendix~\nameref{track2}.

\section{Ablation Study}
This section validates two questions: \textbf{(i)} which components are necessary for safe on-policy shortening, and \textbf{(ii)} how sensitive \our{} is to the two key tolerances that control \textsc{SlackBand} and \textsc{StableSwitch}. Unless stated otherwise, all ablations are conducted on the Logic-RL training setup and evaluated on the \texttt{ppl5} validation protocol used by Logic-RL. We fix the length-reward weight to \(\alpha=1\) throughout to isolate the effects of the gating mechanisms.

\subsection{Component Ablation: Which Gates Matter?}
Recall that \our{} implements a \emph{lazy length penalty} via three gates: \textsc{RightGate} (correctness gating), \textsc{SlackBand} (tolerance band), and \textsc{StableSwitch} (stability-triggered activation). We ablate these components by progressively adding them on top of Standard RL:
\begin{itemize}[leftmargin=15pt]
    \item \textbf{D1 (\textsc{RightGate})}: apply length shaping only on correct trajectories.
    \item \textbf{D1+D2 (\textsc{RightGate}+\textsc{SlackBand})}: additionally penalize only excess length beyond the tolerance band.
    \item \textbf{D1+D3 (\textsc{RightGate}+\textsc{StableSwitch})}: additionally activate length shaping only when accuracy is stable.
    \item \textbf{\our{}}: full method with all three gates.
\end{itemize}
For reference, we also include \textbf{Standard} (no length shaping) and \textbf{Kimi} (direct length reward without lazy gating).

\begin{figure}[htbp]
  \centering
  \subfloat[Response Length]
  {
      \label{fig:alb:subfig1}\includegraphics[width=0.8\linewidth]{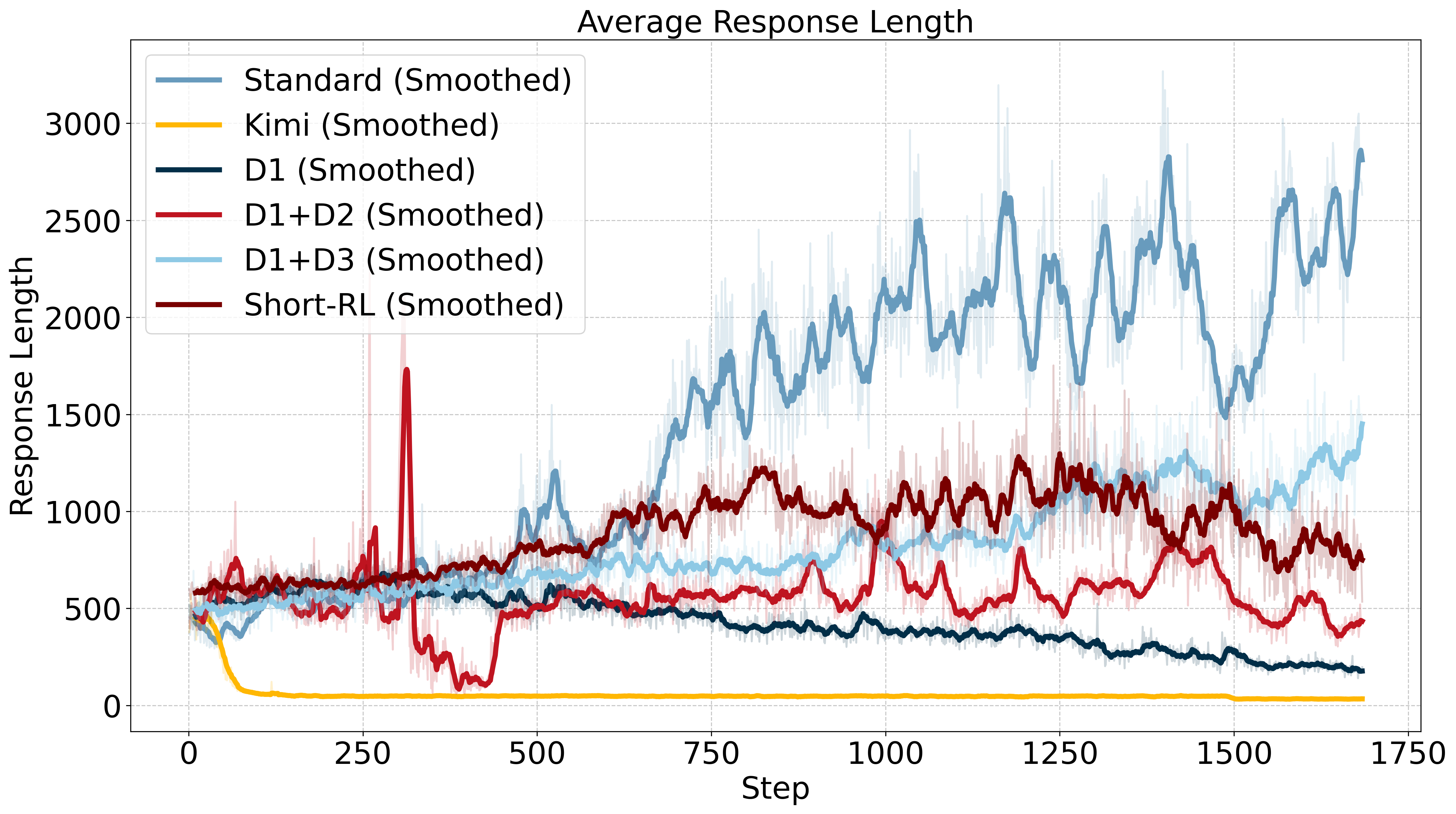}
  }\\
  \subfloat[Accuracy]
  {
      \label{fig:alb:subfig2}\includegraphics[width=0.8\linewidth]{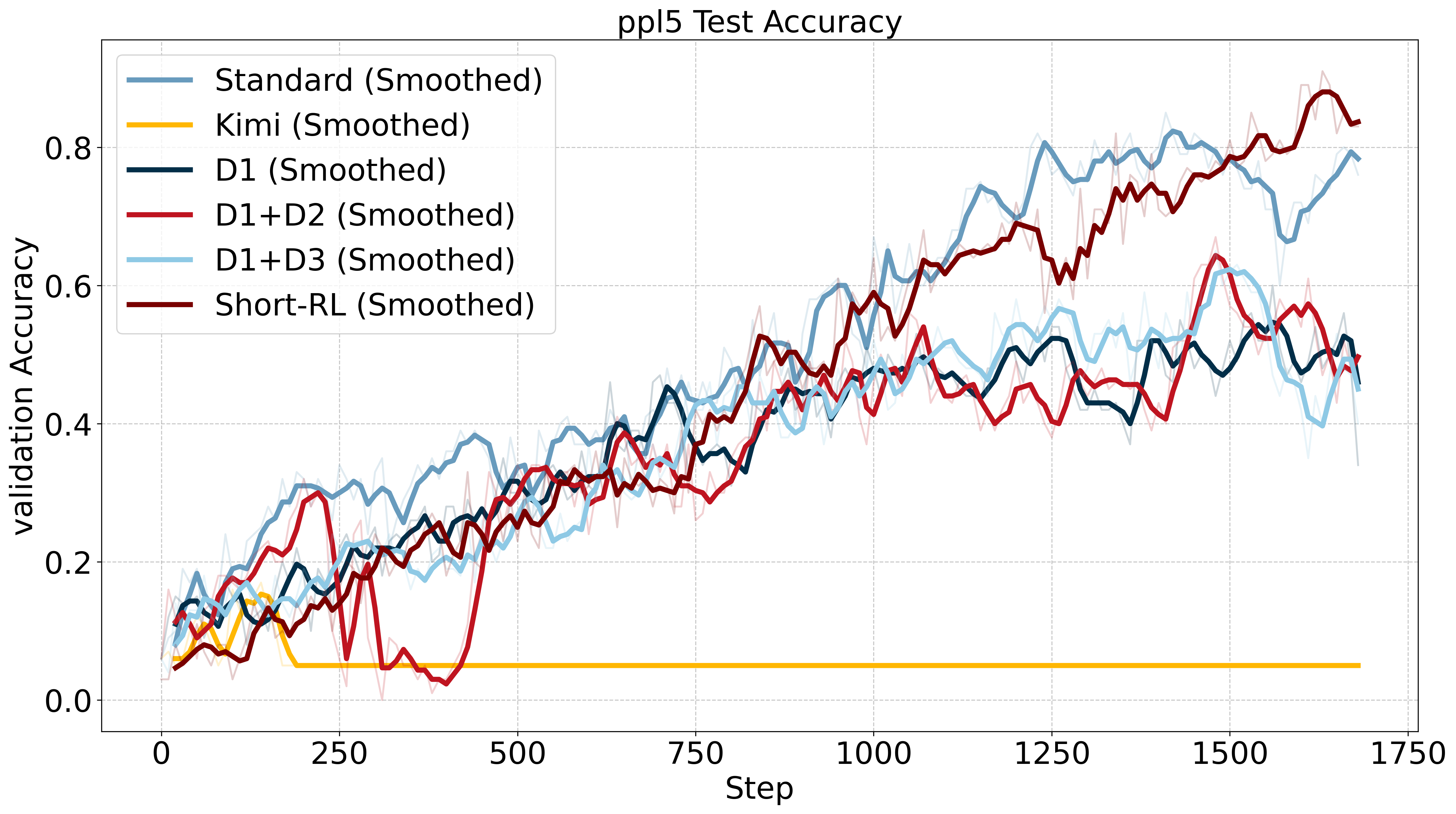}
      \label{fig:alb-b}
  }
  \caption{Component ablation on Logic-RL. Standard RL yields long trajectories; direct Kimi shaping collapses length early. Adding lazy gates progressively stabilizes shortening, and \our{} achieves the best accuracy--length trade-off.}
  \label{fig:alb}
\end{figure}

Figure~\ref{fig:alb} illustrates the training curves for response length and validation accuracy. The results align with our design principle that auxiliary-trajectory regularization should be lazy.
\textbf{First,} Standard RL produces increasingly long trajectories (Figure~\ref{fig:alb:subfig1}), reflecting the well-known tendency of LRMs to lengthen reasoning during on-policy training.
\textbf{Second,} the direct Kimi reward collapses trajectories to very short outputs early, a signature of reward hacking that harms exploration and yields unstable accuracy (Figure~\ref{fig:alb:subfig2}).

In contrast, the proposed gates yield progressively safer shortening.
\textbf{D1} already prevents penalizing incorrect exploratory rollouts and therefore avoids the most aggressive collapse.
Adding \textbf{SlackBand} (\textbf{D1+D2}) further stabilizes behavior by avoiding over-optimization once the model reaches an acceptable concise region.
Adding \textbf{StableSwitch} (\textbf{D1+D3}) improves robustness by turning off length pressure when accuracy dips, preventing length optimization from competing with competence acquisition.
Finally, \textbf{\our{}} combines both safeguards and achieves the best overall trade-off: it shortens trajectories without collapse and attains the highest validation accuracy.

\paragraph{Remark on evaluation protocol.}
Following Logic-RL practice, the \texttt{ppl5} set here is used as a validation set for tracking curves and may differ from the final evaluation split used in Table~\ref{tab:logicrl}. This does not affect the qualitative conclusions of the ablation.

\subsection{Sensitivity: Length Tolerance vs.\ Accuracy Tolerance}
\begin{figure}[htbp]
  \centering
  \subfloat[]
  {
      \label{fig:al:subfig1}\includegraphics[width=0.7\linewidth]{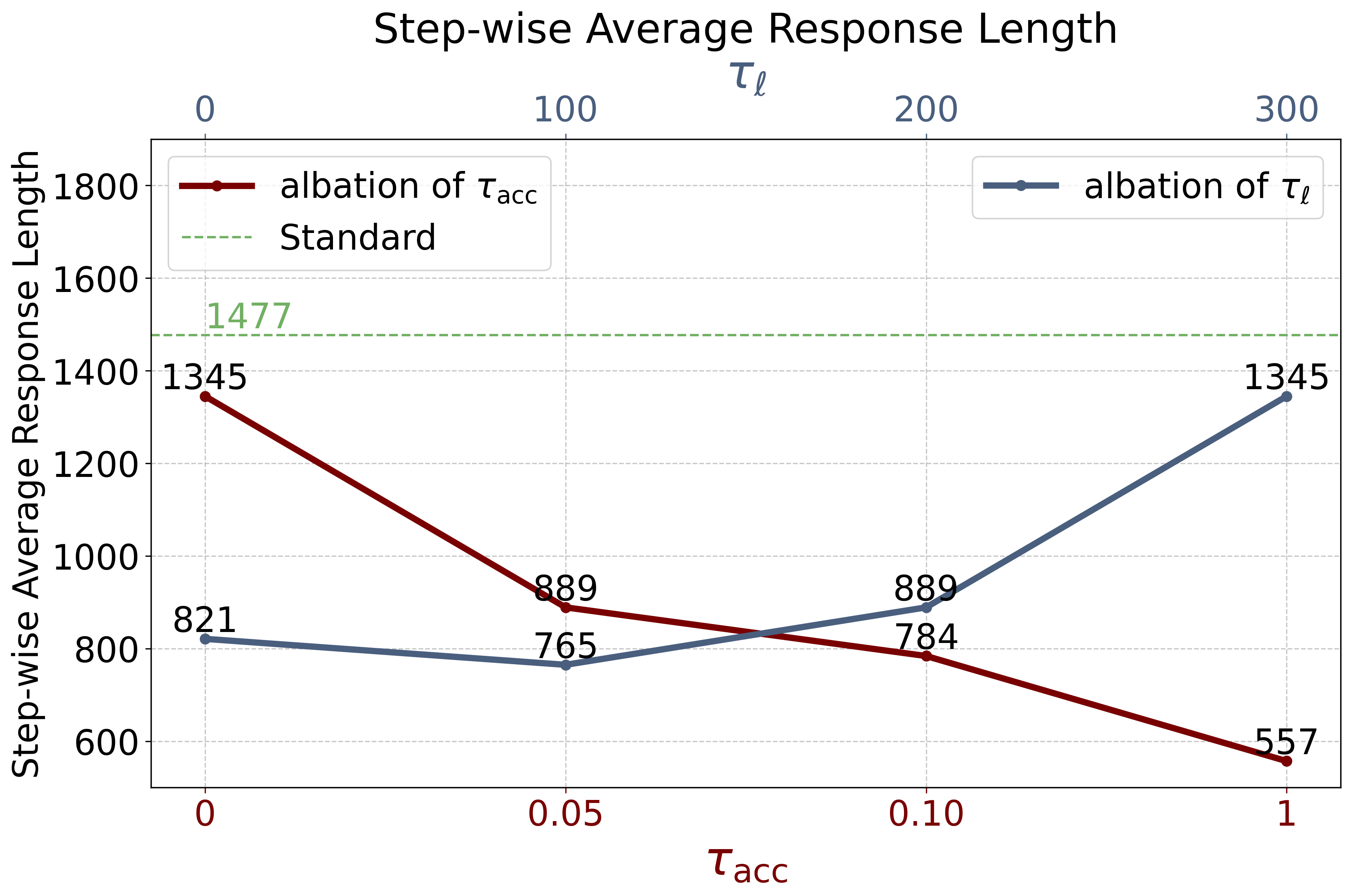}
  }\\
  \subfloat[]
  {
      \label{fig:al:subfig2}\includegraphics[width=0.7\linewidth]{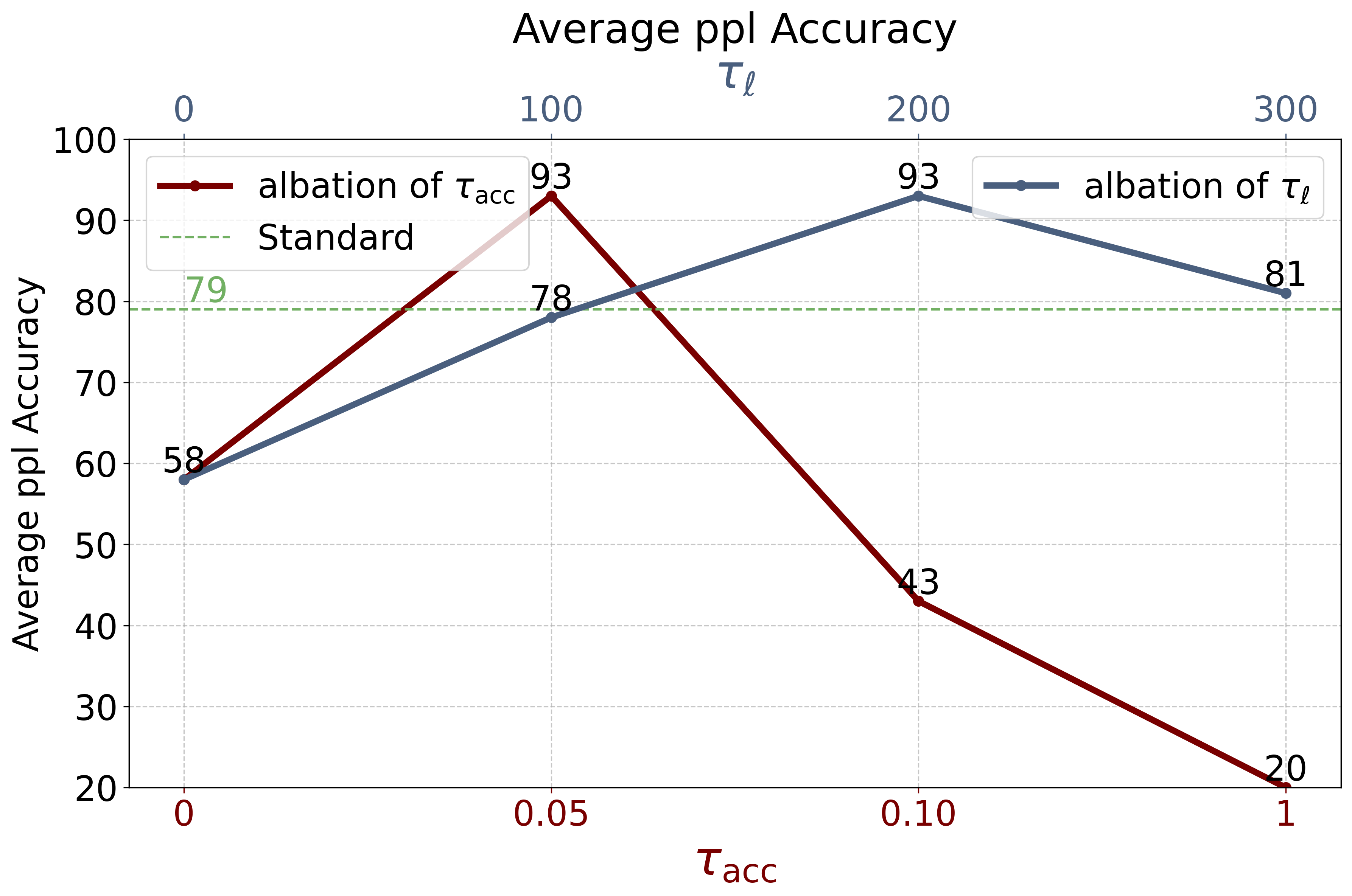}
  }
  \caption{Sensitivity to length tolerance \(\tau_{l}\) (upper x-axis) and accuracy tolerance \(\tau_{\text{acc}}\) (lower x-axis). Smaller \(\tau_{l}\) or larger \(\tau_{\text{acc}}\) increases shortening pressure, which can harm accuracy if overly aggressive.}
  \label{fig:albmerge}
\end{figure}

We next study sensitivity to the two hyperparameters that implement the ``What'' and ``When'' aspects of laziness: the length tolerance \(\tau_{l}\) (for \textsc{SlackBand}) and the accuracy tolerance \(\tau_{\text{acc}}\) (for \textsc{StableSwitch}). Figure~\ref{fig:albmerge} summarizes the resulting average accuracy and step-averaged response length (both averaged over the ppl tasks).

\paragraph{Varying the length tolerance \(\tau_{l}\) (\textsc{SlackBand}).}
We vary \(\tau_{l}\in\{0,100,200,300\}\) while fixing \(\tau_{\text{acc}}=0.05\).
When \(\tau_{l}=0\), the method degenerates toward a strict linear preference for the shortest correct trajectory, which yields the shortest responses but degrades performance---consistent with our claim that overly aggressive length pressure can suppress useful reasoning steps.
For \(\tau_{l}\in\{100,200,300\}\), performance remains stable while response length increases moderately with \(\tau_{l}\). Overall, the method is not overly sensitive to \(\tau_{l}\), and \(\tau_{l}\approx 200\) provides a strong balance in this setting.

\paragraph{Varying the accuracy tolerance \(\tau_{\text{acc}}\) (\textsc{StableSwitch}).}
We vary \(\tau_{\text{acc}}\in\{0,0.05,0.10,1.0\}\) while fixing \(\tau_{l}=200\).
This parameter directly controls how early and how frequently length shaping is activated. Larger \(\tau_{\text{acc}}\) makes activation denser (closer to always-on), which yields shorter trajectories but can degrade accuracy by reintroducing the early-training conflict between length optimization and competence acquisition. In contrast, \(\tau_{\text{acc}}\approx 0.05\) provides a stable and effective operating point, enabling shortening primarily after training becomes sufficiently stable.

\section{Conclusion}
We study how to shorten reasoning trajectories in long reasoning models during \emph{on-policy} rule-based RL, where rollout tokens dominate both training cost and learning dynamics. Our key perspective is that response length is an \emph{auxiliary} trajectory property: correctness defines success, while brevity is a preference among successful trajectories. This motivates \our{}, a \emph{lazy length penalty} that shortens \emph{after the model is right}---it applies length shaping only on correct trajectories (\textsc{RightGate}), only penalizes excess length beyond a tolerance band (\textsc{SlackBand}), and activates only when training accuracy is stable (\textsc{StableSwitch}).

Across logic reasoning and three math RL pipelines, \our{} consistently reduces \textbf{Training (step-avg)} response length (a proxy for training-time rollout token cost) while maintaining or improving accuracy. Compared with post-RL shortening (e.g., Kimi (post)), our on-policy design reduces not only inference length but also the token cost incurred \emph{during} RL training.

\section{Limitations}
\paragraph{Scope of applicability.}
\our{} targets settings where the model produces an explicit multi-step reasoning trajectory and correctness can be reliably assessed with a rule-based signal (e.g., math and logic). In such tasks, excessive length often reflects redundancy, making trajectory shortening both meaningful and measurable. For open-ended generation tasks (e.g., creative writing or dialogue), there may be no clear notion of ``correct'' answers or minimal-length successful trajectories; in these cases, imposing a length preference can be misaligned with the objective or reduce desirable stylistic variation.

\paragraph{Dependence on reward reliability.}
Our lazy gates rely on a reasonably accurate correctness signal (\(c_i\)) to determine which trajectories are eligible for length shaping. If the rule-based reward is noisy or can be exploited, the system may apply length pressure to trajectories that are not truly correct, potentially biasing learning. While this concern is shared by many rule-based RL pipelines, it becomes more salient when auxiliary trajectory shaping is introduced.

\clearpage

\bibliography{iclr2026_conference}

\appendix
\section{Appendix}

\begin{table*}[t]
  \setlength{\tabcolsep}{3.0pt}
  \small
  \centering
  \scalebox{0.9}{
  \begin{tabular}{lcccc}
  \toprule
  \bf Setting  & \bf Logic-RL & \bf DeepScaleR   & \bf Open-Reasoner-Zero  & \bf SimpleRL-Reason \\
  \midrule
 learning rate  & 1e-6 & 1e-6 & 5e-7 & 5e-7   \\
  batch size  &8 & 128 & 64 & 16   \\
   ppo\_mini\_batch\_size  &32 & 64 & 256 & 64   \\
    ppo\_micro\_batch\_size  &8 & 32 & 64 & 2   \\
     rollout\_n   & 8 & 8 & 8 & 8   \\
     temperature   & 0.7 & 0.6 & 1.0 & 1.0   \\
     kl\_loss\_coef & 0.001 & 0.001  & 0.001 &0.0001\\
     epochs & 3 & 3  & 1 &3 \\
     max\_response\_length & 4096 & 8192 & 4096 & 8192\\
     algorithm & reinforce++ & grpo & grpo & grpo \\
     \(\tau_{l}\)  & 200 & 100 & 100 & 50 \\
     \(\tau_{\text{acc}}\) & 0.05 & 0.05 & 0.02 & 0.05 \\
     \(\alpha\) & 1 & 1 & 1 & 1 \\
     Model & Qwen2.5-7B  & DeepSeek Distill Qwen-1.5B & Qwen2.5-7B  & Qwen2.5-7B \\
  \bottomrule
  \end{tabular}
  }
\caption{Training details.}
\label{tab:detail}
\end{table*}

\subsection{Additional Experiments}
\label{addition}
\begin{figure}[htbp]
  \centering
  \begin{minipage}[t]{0.8\linewidth}
  \subfloat[Open-Reasoner-Zero]
  {\label{fig:lcmc:subfig1}\includegraphics[width=0.9\textwidth]{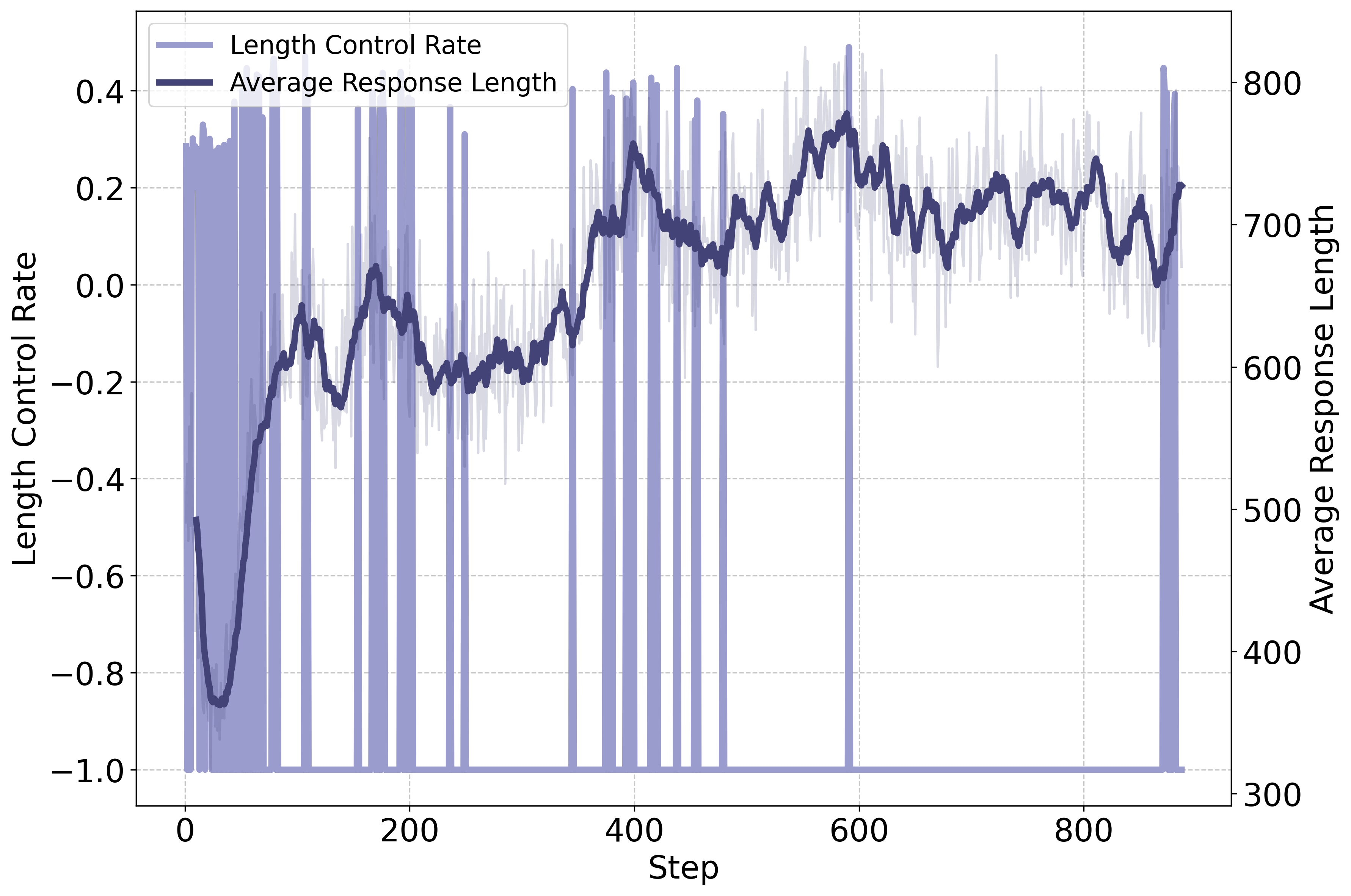}}
  \end{minipage}\\
  \begin{minipage}[t]{0.8\linewidth}
  \subfloat[SimpleRL-Reason]
  {\label{fig:lcmc:subfig2}\includegraphics[width=0.9\textwidth]{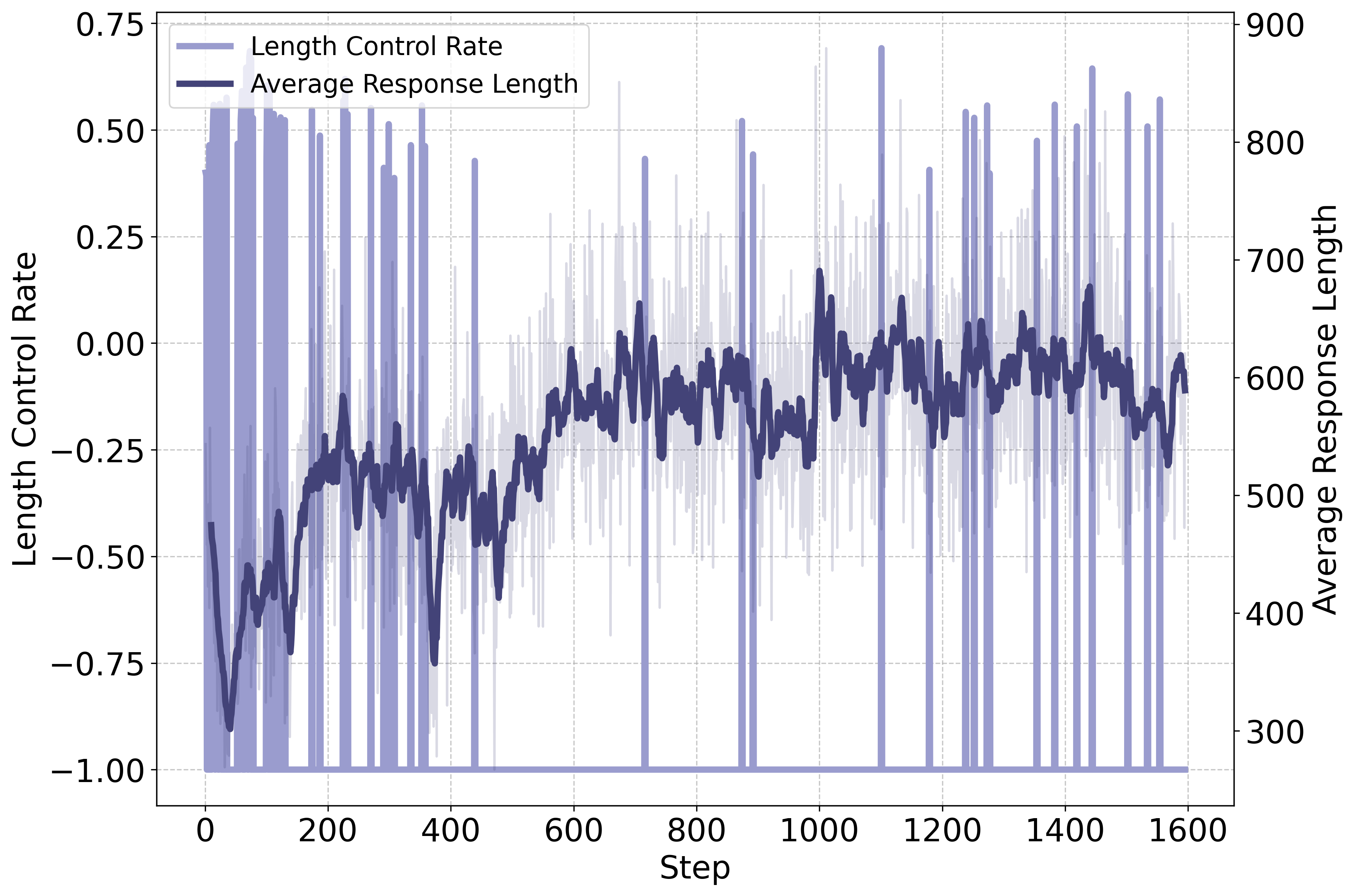}}
  \end{minipage}
  \caption{Visualization of the length control rate during training.}
  \label{fig:addlen_control}
\end{figure}

\subsubsection{Track the Length Reward}
\label{track2}
We also track the metric defined in Section \nameref{track} in Figure~\ref{fig:addlen_control} (Open-Reasoner-Zero and SimpleRL-Reason).

\begin{figure}[htbp]
  \centering
  \subfloat[]
  {
      \includegraphics[width=0.8\linewidth]{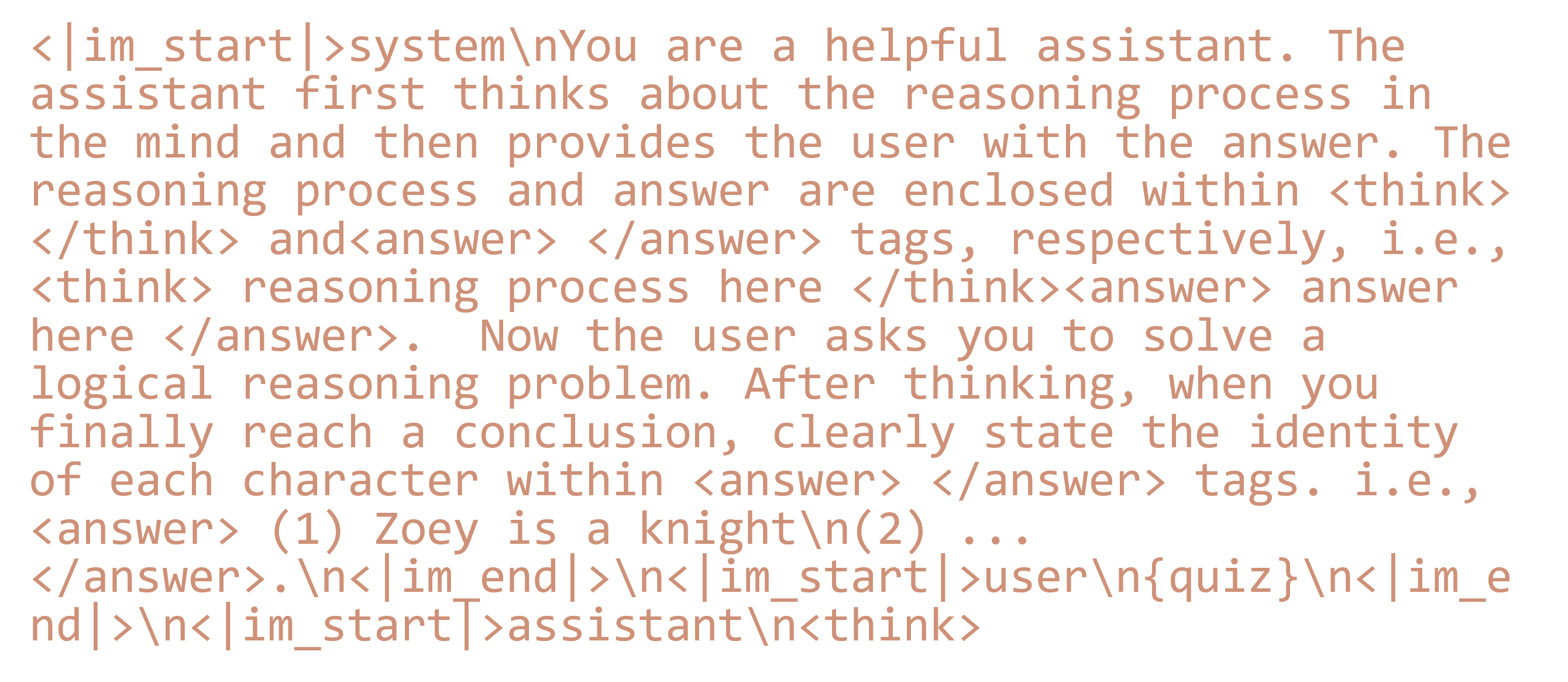}
      \label{fig:template:subfig1}
  }\\
  \subfloat[]
  {
      \includegraphics[width=0.8\linewidth]{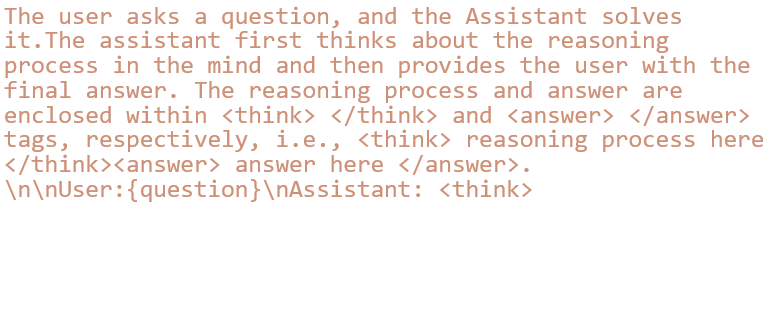}
       \label{fig:template:subfig2}
  }
  \caption{The prompt template for Logic-RL and Math-RL.}
  \label{fig:template}
\end{figure}

\subsection{Training Details}
\label{details}
Our experiments were conducted using a compute node equipped with 8 NVIDIA H100 GPUs. The CUDA version we use is 12.3.

\subsubsection{Logic-RL Training and Evaluation Details}
The training and evaluation prompt template (Figure~\ref{fig:template:subfig1}) used in Logic-RL remains the same as in the original GitHub project. The training hyperparameters are listed in Table~\ref{tab:detail}.
During evaluation, we directly use the code from Logic-RL, which applies a temperature of 1.0 and top\_p=1.0 for logic tasks, and a temperature of 0.8 with top\_p=0.95 for math tasks.

\subsubsection{Training and Evaluation Details for Math}
The training and evaluation prompt template for three math settings is shown in Figure~\ref{fig:template:subfig2}. The training hyperparameters are listed in Table~\ref{tab:detail}.
During evaluation, we directly use the code from DeepScaleR, which employs a temperature of 1.0.

\subsubsection{Reward Details}
\label{app:format}
In all the math experiments, the standard task reward employs a format and outcome-based scheme:
\begin{equation}
\begin{aligned}
R_{\text{task}} & =
\begin{cases}
\, 3, \, &\text{format correct and answer correct}\\
 -0.5,  \, &\text{format correct and answer wrong} \\
- 3, \, &\text{format wrong.}
\end{cases}
\end{aligned}
\end{equation}
In Logic-RL experiments, we directly use their original standard reward design.

\end{document}

%% file: math_commands.tex
\usepackage{amsmath,amsfonts,bm}

\def\eqref#1{equation~\ref{#1}}

\def\1{\bm{1}}

\DeclareMathAlphabet{\mathsfit}{\encodingdefault}{\sfdefault}{m}{sl}
\SetMathAlphabet{\mathsfit}{bold}{\encodingdefault}{\sfdefault}{bx}{n}



%% file: aaai2026.bib
@misc{openai2024openaio1card,
      title={OpenAI o1 System Card}, 
      author={OpenAI and : and Aaron Jaech and Adam Kalai and Adam Lerer and Adam Richardson and Ahmed El-Kishky and Aiden Low and Alec Helyar and Aleksander Madry and Alex Beutel and Alex Carney and Alex Iftimie and Alex Karpenko and Alex Tachard Passos and Alexander Neitz and Alexander Prokofiev and Alexander Wei and Allison Tam and Ally Bennett and Ananya Kumar and Andre Saraiva and Andrea Vallone and Andrew Duberstein and Andrew Kondrich and Andrey Mishchenko and Andy Applebaum and Angela Jiang and Ashvin Nair and Barret Zoph and Behrooz Ghorbani and Ben Rossen and Benjamin Sokolowsky and Boaz Barak and Bob McGrew and Borys Minaiev and Botao Hao and Bowen Baker and Brandon Houghton and Brandon McKinzie and Brydon Eastman and Camillo Lugaresi and Cary Bassin and Cary Hudson and Chak Ming Li and Charles de Bourcy and Chelsea Voss and Chen Shen and Chong Zhang and Chris Koch and Chris Orsinger and Christopher Hesse and Claudia Fischer and Clive Chan and Dan Roberts and Daniel Kappler and Daniel Levy and Daniel Selsam and David Dohan and David Farhi and David Mely and David Robinson and Dimitris Tsipras and Doug Li and Dragos Oprica and Eben Freeman and Eddie Zhang and Edmund Wong and Elizabeth Proehl and Enoch Cheung and Eric Mitchell and Eric Wallace and Erik Ritter and Evan Mays and Fan Wang and Felipe Petroski Such and Filippo Raso and Florencia Leoni and Foivos Tsimpourlas and Francis Song and Fred von Lohmann and Freddie Sulit and Geoff Salmon and Giambattista Parascandolo and Gildas Chabot and Grace Zhao and Greg Brockman and Guillaume Leclerc and Hadi Salman and Haiming Bao and Hao Sheng and Hart Andrin and Hessam Bagherinezhad and Hongyu Ren and Hunter Lightman and Hyung Won Chung and Ian Kivlichan and Ian O'Connell and Ian Osband and Ignasi Clavera Gilaberte and Ilge Akkaya and Ilya Kostrikov and Ilya Sutskever and Irina Kofman and Jakub Pachocki and James Lennon and Jason Wei and Jean Harb and Jerry Twore and Jiacheng Feng and Jiahui Yu and Jiayi Weng and Jie Tang and Jieqi Yu and Joaquin Quiñonero Candela and Joe Palermo and Joel Parish and Johannes Heidecke and John Hallman and John Rizzo and Jonathan Gordon and Jonathan Uesato and Jonathan Ward and Joost Huizinga and Julie Wang and Kai Chen and Kai Xiao and Karan Singhal and Karina Nguyen and Karl Cobbe and Katy Shi and Kayla Wood and Kendra Rimbach and Keren Gu-Lemberg and Kevin Liu and Kevin Lu and Kevin Stone and Kevin Yu and Lama Ahmad and Lauren Yang and Leo Liu and Leon Maksin and Leyton Ho and Liam Fedus and Lilian Weng and Linden Li and Lindsay McCallum and Lindsey Held and Lorenz Kuhn and Lukas Kondraciuk and Lukasz Kaiser and Luke Metz and Madelaine Boyd and Maja Trebacz and Manas Joglekar and Mark Chen and Marko Tintor and Mason Meyer and Matt Jones and Matt Kaufer and Max Schwarzer and Meghan Shah and Mehmet Yatbaz and Melody Y. Guan and Mengyuan Xu and Mengyuan Yan and Mia Glaese and Mianna Chen and Michael Lampe and Michael Malek and Michele Wang and Michelle Fradin and Mike McClay and Mikhail Pavlov and Miles Wang and Mingxuan Wang and Mira Murati and Mo Bavarian and Mostafa Rohaninejad and Nat McAleese and Neil Chowdhury and Neil Chowdhury and Nick Ryder and Nikolas Tezak and Noam Brown and Ofir Nachum and Oleg Boiko and Oleg Murk and Olivia Watkins and Patrick Chao and Paul Ashbourne and Pavel Izmailov and Peter Zhokhov and Rachel Dias and Rahul Arora and Randall Lin and Rapha Gontijo Lopes and Raz Gaon and Reah Miyara and Reimar Leike and Renny Hwang and Rhythm Garg and Robin Brown and Roshan James and Rui Shu and Ryan Cheu and Ryan Greene and Saachi Jain and Sam Altman and Sam Toizer and Sam Toyer and Samuel Miserendino and Sandhini Agarwal and Santiago Hernandez and Sasha Baker and Scott McKinney and Scottie Yan and Shengjia Zhao and Shengli Hu and Shibani Santurkar and Shraman Ray Chaudhuri and Shuyuan Zhang and Siyuan Fu and Spencer Papay and Steph Lin and Suchir Balaji and Suvansh Sanjeev and Szymon Sidor and Tal Broda and Aidan Clark and Tao Wang and Taylor Gordon and Ted Sanders and Tejal Patwardhan and Thibault Sottiaux and Thomas Degry and Thomas Dimson and Tianhao Zheng and Timur Garipov and Tom Stasi and Trapit Bansal and Trevor Creech and Troy Peterson and Tyna Eloundou and Valerie Qi and Vineet Kosaraju and Vinnie Monaco and Vitchyr Pong and Vlad Fomenko and Weiyi Zheng and Wenda Zhou and Wes McCabe and Wojciech Zaremba and Yann Dubois and Yinghai Lu and Yining Chen and Young Cha and Yu Bai and Yuchen He and Yuchen Zhang and Yunyun Wang and Zheng Shao and Zhuohan Li},
      year={2024},
      eprint={2412.16720},
      archivePrefix={arXiv},
      primaryClass={cs.AI},
      url={https://arxiv.org/abs/2412.16720}, 
}

@misc{deepseekai2025deepseekr1incentivizingreasoningcapability,
      title={DeepSeek-R1: Incentivizing Reasoning Capability in LLMs via Reinforcement Learning}, 
      author={DeepSeek-AI and Daya Guo and Dejian Yang and Haowei Zhang and Junxiao Song and Ruoyu Zhang and Runxin Xu and Qihao Zhu and Shirong Ma and Peiyi Wang and Xiao Bi and Xiaokang Zhang and Xingkai Yu and Yu Wu and Z. F. Wu and Zhibin Gou and Zhihong Shao and Zhuoshu Li and Ziyi Gao and Aixin Liu and Bing Xue and Bingxuan Wang and Bochao Wu and Bei Feng and Chengda Lu and Chenggang Zhao and Chengqi Deng and Chenyu Zhang and Chong Ruan and Damai Dai and Deli Chen and Dongjie Ji and Erhang Li and Fangyun Lin and Fucong Dai and Fuli Luo and Guangbo Hao and Guanting Chen and Guowei Li and H. Zhang and Han Bao and Hanwei Xu and Haocheng Wang and Honghui Ding and Huajian Xin and Huazuo Gao and Hui Qu and Hui Li and Jianzhong Guo and Jiashi Li and Jiawei Wang and Jingchang Chen and Jingyang Yuan and Junjie Qiu and Junlong Li and J. L. Cai and Jiaqi Ni and Jian Liang and Jin Chen and Kai Dong and Kai Hu and Kaige Gao and Kang Guan and Kexin Huang and Kuai Yu and Lean Wang and Lecong Zhang and Liang Zhao and Litong Wang and Liyue Zhang and Lei Xu and Leyi Xia and Mingchuan Zhang and Minghua Zhang and Minghui Tang and Meng Li and Miaojun Wang and Mingming Li and Ning Tian and Panpan Huang and Peng Zhang and Qiancheng Wang and Qinyu Chen and Qiushi Du and Ruiqi Ge and Ruisong Zhang and Ruizhe Pan and Runji Wang and R. J. Chen and R. L. Jin and Ruyi Chen and Shanghao Lu and Shangyan Zhou and Shanhuang Chen and Shengfeng Ye and Shiyu Wang and Shuiping Yu and Shunfeng Zhou and Shuting Pan and S. S. Li and Shuang Zhou and Shaoqing Wu and Shengfeng Ye and Tao Yun and Tian Pei and Tianyu Sun and T. Wang and Wangding Zeng and Wanjia Zhao and Wen Liu and Wenfeng Liang and Wenjun Gao and Wenqin Yu and Wentao Zhang and W. L. Xiao and Wei An and Xiaodong Liu and Xiaohan Wang and Xiaokang Chen and Xiaotao Nie and Xin Cheng and Xin Liu and Xin Xie and Xingchao Liu and Xinyu Yang and Xinyuan Li and Xuecheng Su and Xuheng Lin and X. Q. Li and Xiangyue Jin and Xiaojin Shen and Xiaosha Chen and Xiaowen Sun and Xiaoxiang Wang and Xinnan Song and Xinyi Zhou and Xianzu Wang and Xinxia Shan and Y. K. Li and Y. Q. Wang and Y. X. Wei and Yang Zhang and Yanhong Xu and Yao Li and Yao Zhao and Yaofeng Sun and Yaohui Wang and Yi Yu and Yichao Zhang and Yifan Shi and Yiliang Xiong and Ying He and Yishi Piao and Yisong Wang and Yixuan Tan and Yiyang Ma and Yiyuan Liu and Yongqiang Guo and Yuan Ou and Yuduan Wang and Yue Gong and Yuheng Zou and Yujia He and Yunfan Xiong and Yuxiang Luo and Yuxiang You and Yuxuan Liu and Yuyang Zhou and Y. X. Zhu and Yanhong Xu and Yanping Huang and Yaohui Li and Yi Zheng and Yuchen Zhu and Yunxian Ma and Ying Tang and Yukun Zha and Yuting Yan and Z. Z. Ren and Zehui Ren and Zhangli Sha and Zhe Fu and Zhean Xu and Zhenda Xie and Zhengyan Zhang and Zhewen Hao and Zhicheng Ma and Zhigang Yan and Zhiyu Wu and Zihui Gu and Zijia Zhu and Zijun Liu and Zilin Li and Ziwei Xie and Ziyang Song and Zizheng Pan and Zhen Huang and Zhipeng Xu and Zhongyu Zhang and Zhen Zhang},
      year={2025},
      eprint={2501.12948},
      archivePrefix={arXiv},
      primaryClass={cs.CL},
      url={https://arxiv.org/abs/2501.12948}, 
}

@misc{xie2025logicrlunleashingllmreasoning,
      title={Logic-RL: Unleashing LLM Reasoning with Rule-Based Reinforcement Learning}, 
      author={Tian Xie and Zitian Gao and Qingnan Ren and Haoming Luo and Yuqian Hong and Bryan Dai and Joey Zhou and Kai Qiu and Zhirong Wu and Chong Luo},
      year={2025},
      eprint={2502.14768},
      archivePrefix={arXiv},
      primaryClass={cs.CL},
      url={https://arxiv.org/abs/2502.14768}, 
}

@article{Xie2024OnMO,
  title={On Memorization of Large Language Models in Logical Reasoning},
  author={Chulin Xie and Yangsibo Huang and Chiyuan Zhang and Da Yu and Xinyun Chen and Bill Yuchen Lin and Bo Li and Badih Ghazi and Ravi Kumar},
  journal={ArXiv},
  year={2024},
  volume={abs/2410.23123},
  url={https://api.semanticscholar.org/CorpusID:273695832}
}

@misc{simplerlmath2025,
      title={SimpleRL-Zoo: Investigating and Taming Zero Reinforcement Learning for Open Base Models in the Wild}, 
      author={Weihao Zeng and Yuzhen Huang and Qian Liu and Wei Liu and Keqing He and Zejun Ma and Junxian He},
      year={2025},
      eprint={2503.18892},
      archivePrefix={arXiv},
      primaryClass={cs.LG},
      url={https://arxiv.org/abs/2503.18892}, 
}

@misc{deepscaler2025,
  title={DeepScaleR: Surpassing O1-Preview with a 1.5B Model by Scaling RL},
  author={Michael Luo and Sijun Tan and Justin Wong and Xiaoxiang Shi and William Y. Tang and Manan Roongta and Colin Cai and Jeffrey Luo and Tianjun Zhang and Li Erran Li and Raluca Ada Popa and Ion Stoica},
  year={2025},
  howpublished={https://pretty-radio-b75.notion.site/DeepScaleR-Surpassing-O1-Preview-with-a-1-5B-Model-by-Scaling-RL-19681902c1468005bed8ca303013a4e2},
  note={Notion Blog},
}

@misc{OpenReasonerZero2025,
  title={Open-Reasoner-Zero: An Open Source Approach to Scaling Reinforcement Learning on the Base Model},
  author={Jingcheng Hu and Yinmin Zhang and Qi Han and Daxin Jiang and Xiangyu Zhang, Heung-Yeung Shum},
  year={2025},
  howpublished={https://github.com/Open-Reasoner-Zero/Open-Reasoner-Zero},
}

@misc{dapo2025,
      title={DAPO: An Open-Source LLM Reinforcement Learning System at Scale}, 
      author={Qiying Yu and Zheng Zhang and Ruofei Zhu and Yufeng Yuan and Xiaochen Zuo and Yu Yue and Tiantian Fan and Gaohong Liu and Lingjun Liu and Xin Liu and Haibin Lin and Zhiqi Lin and Bole Ma and Guangming Sheng and Yuxuan Tong and Chi Zhang and Mofan Zhang and Wang Zhang and Hang Zhu and Jinhua Zhu and Jiaze Chen and Jiangjie Chen and Chengyi Wang and Hongli Yu and Weinan Dai and Yuxuan Song and Xiangpeng Wei and Hao Zhou and Jingjing Liu and Wei-Ying Ma and Ya-Qin Zhang and Lin Yan and Mu Qiao and Yonghui Wu and Mingxuan Wang},
      year={2025},
      eprint={2503.14476},
      archivePrefix={arXiv},
      primaryClass={cs.LG},
      url={https://arxiv.org/abs/2503.14476}, 
}

@misc{sui2025stopoverthinkingsurveyefficient,
      title={Stop Overthinking: A Survey on Efficient Reasoning for Large Language Models}, 
      author={Yang Sui and Yu-Neng Chuang and Guanchu Wang and Jiamu Zhang and Tianyi Zhang and Jiayi Yuan and Hongyi Liu and Andrew Wen and Shaochen Zhong and Hanjie Chen and Xia Hu},
      year={2025},
      eprint={2503.16419},
      archivePrefix={arXiv},
      primaryClass={cs.CL},
      url={https://arxiv.org/abs/2503.16419}, 
}

@misc{tokenskip2025,
      title={TokenSkip: Controllable Chain-of-Thought Compression in LLMs}, 
      author={Heming Xia and Yongqi Li and Chak Tou Leong and Wenjie Wang and Wenjie Li},
      year={2025},
      eprint={2502.12067},
      archivePrefix={arXiv},
      primaryClass={cs.CL},
      url={https://arxiv.org/abs/2502.12067}, 
}

@misc{C3ot2025,
      title={C3oT: Generating Shorter Chain-of-Thought without Compromising Effectiveness}, 
      author={Yu Kang and Xianghui Sun and Liangyu Chen and Wei Zou},
      year={2024},
      eprint={2412.11664},
      archivePrefix={arXiv},
      primaryClass={cs.CL},
      url={https://arxiv.org/abs/2412.11664}, 
}

@misc{cotValue2025,
      title={CoT-Valve: Length-Compressible Chain-of-Thought Tuning}, 
      author={Xinyin Ma and Guangnian Wan and Runpeng Yu and Gongfan Fang and Xinchao Wang},
      year={2025},
      eprint={2502.09601},
      archivePrefix={arXiv},
      primaryClass={cs.AI},
      url={https://arxiv.org/abs/2502.09601}, 
}

@misc{selftrain2025,
      title={Self-Training Elicits Concise Reasoning in Large Language Models}, 
      author={Tergel Munkhbat and Namgyu Ho and Seo Hyun Kim and Yongjin Yang and Yujin Kim and Se-Young Yun},
      year={2025},
      eprint={2502.20122},
      archivePrefix={arXiv},
      primaryClass={cs.CL},
      url={https://arxiv.org/abs/2502.20122}, 
}

@misc{distilling2to12024,
      title={Distilling System 2 into System 1}, 
      author={Ping Yu and Jing Xu and Jason Weston and Ilia Kulikov},
      year={2024},
      eprint={2407.06023},
      archivePrefix={arXiv},
      primaryClass={cs.CL},
      url={https://arxiv.org/abs/2407.06023}, 
}

@misc{skipsteps2024,
      title={Can Language Models Learn to Skip Steps?}, 
      author={Tengxiao Liu and Qipeng Guo and Xiangkun Hu and Cheng Jiayang and Yue Zhang and Xipeng Qiu and Zheng Zhang},
      year={2024},
      eprint={2411.01855},
      archivePrefix={arXiv},
      primaryClass={cs.CL},
      url={https://arxiv.org/abs/2411.01855}, 
}

@misc{stepre2025,
      title={Stepwise Perplexity-Guided Refinement for Efficient Chain-of-Thought Reasoning in Large Language Models}, 
      author={Yingqian Cui and Pengfei He and Jingying Zeng and Hui Liu and Xianfeng Tang and Zhenwei Dai and Yan Han and Chen Luo and Jing Huang and Zhen Li and Suhang Wang and Yue Xing and Jiliang Tang and Qi He},
      year={2025},
      eprint={2502.13260},
      archivePrefix={arXiv},
      primaryClass={cs.CL},
      url={https://arxiv.org/abs/2502.13260}, 
}

@misc{o1-pruner2025,
      title={O1-Pruner: Length-Harmonizing Fine-Tuning for O1-Like Reasoning Pruning}, 
      author={Haotian Luo and Li Shen and Haiying He and Yibo Wang and Shiwei Liu and Wei Li and Naiqiang Tan and Xiaochun Cao and Dacheng Tao},
      year={2025},
      eprint={2501.12570},
      archivePrefix={arXiv},
      primaryClass={cs.CL},
      url={https://arxiv.org/abs/2501.12570}, 
}

@misc{DAST2025,
      title={DAST: Difficulty-Adaptive Slow-Thinking for Large Reasoning Models}, 
      author={Yi Shen and Jian Zhang and Jieyun Huang and Shuming Shi and Wenjing Zhang and Jiangze Yan and Ning Wang and Kai Wang and Shiguo Lian},
      year={2025},
      eprint={2503.04472},
      archivePrefix={arXiv},
      primaryClass={cs.LG},
      url={https://arxiv.org/abs/2503.04472}, 
}

@misc{kimi2025,
      title={Kimi k1.5: Scaling Reinforcement Learning with LLMs}, 
      author={Kimi Team and Angang Du and Bofei Gao and Bowei Xing and Changjiu Jiang and Cheng Chen and Cheng Li and Chenjun Xiao and Chenzhuang Du and Chonghua Liao and Chuning Tang and Congcong Wang and Dehao Zhang and Enming Yuan and Enzhe Lu and Fengxiang Tang and Flood Sung and Guangda Wei and Guokun Lai and Haiqing Guo and Han Zhu and Hao Ding and Hao Hu and Hao Yang and Hao Zhang and Haotian Yao and Haotian Zhao and Haoyu Lu and Haoze Li and Haozhen Yu and Hongcheng Gao and Huabin Zheng and Huan Yuan and Jia Chen and Jianhang Guo and Jianlin Su and Jianzhou Wang and Jie Zhao and Jin Zhang and Jingyuan Liu and Junjie Yan and Junyan Wu and Lidong Shi and Ling Ye and Longhui Yu and Mengnan Dong and Neo Zhang and Ningchen Ma and Qiwei Pan and Qucheng Gong and Shaowei Liu and Shengling Ma and Shupeng Wei and Sihan Cao and Siying Huang and Tao Jiang and Weihao Gao and Weimin Xiong and Weiran He and Weixiao Huang and Wenhao Wu and Wenyang He and Xianghui Wei and Xianqing Jia and Xingzhe Wu and Xinran Xu and Xinxing Zu and Xinyu Zhou and Xuehai Pan and Y. Charles and Yang Li and Yangyang Hu and Yangyang Liu and Yanru Chen and Yejie Wang and Yibo Liu and Yidao Qin and Yifeng Liu and Ying Yang and Yiping Bao and Yulun Du and Yuxin Wu and Yuzhi Wang and Zaida Zhou and Zhaoji Wang and Zhaowei Li and Zhen Zhu and Zheng Zhang and Zhexu Wang and Zhilin Yang and Zhiqi Huang and Zihao Huang and Ziyao Xu and Zonghan Yang},
      year={2025},
      eprint={2501.12599},
      archivePrefix={arXiv},
      primaryClass={cs.AI},
      url={https://arxiv.org/abs/2501.12599}, 
}

@misc{qwen2025,
      title={Qwen2.5 Technical Report}, 
      author={Qwen and : and An Yang and Baosong Yang and Beichen Zhang and Binyuan Hui and Bo Zheng and Bowen Yu and Chengyuan Li and Dayiheng Liu and Fei Huang and Haoran Wei and Huan Lin and Jian Yang and Jianhong Tu and Jianwei Zhang and Jianxin Yang and Jiaxi Yang and Jingren Zhou and Junyang Lin and Kai Dang and Keming Lu and Keqin Bao and Kexin Yang and Le Yu and Mei Li and Mingfeng Xue and Pei Zhang and Qin Zhu and Rui Men and Runji Lin and Tianhao Li and Tianyi Tang and Tingyu Xia and Xingzhang Ren and Xuancheng Ren and Yang Fan and Yang Su and Yichang Zhang and Yu Wan and Yuqiong Liu and Zeyu Cui and Zhenru Zhang and Zihan Qiu},
      year={2025},
      eprint={2412.15115},
      archivePrefix={arXiv},
      primaryClass={cs.CL},
      url={https://arxiv.org/abs/2412.15115}, 
}

@misc{overthinking25,
      title={Do NOT Think That Much for 2+3=? On the Overthinking of o1-Like LLMs}, 
      author={Xingyu Chen and Jiahao Xu and Tian Liang and Zhiwei He and Jianhui Pang and Dian Yu and Linfeng Song and Qiuzhi Liu and Mengfei Zhou and Zhuosheng Zhang and Rui Wang and Zhaopeng Tu and Haitao Mi and Dong Yu},
      year={2025},
      eprint={2412.21187},
      archivePrefix={arXiv},
      primaryClass={cs.CL},
      url={https://arxiv.org/abs/2412.21187}, 
}

@misc{2025tokenbudget,
      title={Token-Budget-Aware LLM Reasoning}, 
      author={Tingxu Han and Zhenting Wang and Chunrong Fang and Shiyu Zhao and Shiqing Ma and Zhenyu Chen},
      year={2025},
      eprint={2412.18547},
      archivePrefix={arXiv},
      primaryClass={cs.CL},
      url={https://arxiv.org/abs/2412.18547}, 
}

@misc{2025chaindraft,
      title={Chain of Draft: Thinking Faster by Writing Less}, 
      author={Silei Xu and Wenhao Xie and Lingxiao Zhao and Pengcheng He},
      year={2025}, 
      eprint={2502.18600},
      archivePrefix={arXiv},
      primaryClass={cs.CL},
      url={https://arxiv.org/abs/2502.18600}, 
}

@inproceedings{2024Concise,
   title={The Benefits of a Concise Chain of Thought on Problem-Solving in Large Language Models},
   url={http://dx.doi.org/10.1109/FLLM63129.2024.10852493},
   DOI={10.1109/fllm63129.2024.10852493},
   booktitle={2024 2nd International Conference on Foundation and Large Language Models (FLLM)},
   publisher={IEEE},
   author={Renze, Matthew and Guven, Erhan},
   year={2024},
   month=nov, pages={476–483} }

@misc{NoThink,
      title={Reasoning Models Can Be Effective Without Thinking}, 
      author={Wenjie Ma and Jingxuan He and Charlie Snell and Tyler Griggs and Sewon Min and Matei Zaharia},
      year={2025},
      eprint={2504.09858},
      archivePrefix={arXiv},
      primaryClass={cs.AI},
      url={https://arxiv.org/abs/2504.09858}, 
}

@article{Arora2025TrainingLM,
  title={Training Language Models to Reason Efficiently},
  author={Daman Arora and Andrea Zanette},
  journal={ArXiv},
  year={2025},
  volume={abs/2502.04463},
  url={https://api.semanticscholar.org/CorpusID:276235717}
}

@misc{modelcollaboration,
      title={Hawkeye:Efficient Reasoning with Model Collaboration}, 
      author={Jianshu She and Zhuohao Li and Zhemin Huang and Qi Li and Peiran Xu and Haonan Li and Qirong Ho},
      year={2025},
      eprint={2504.00424},
      archivePrefix={arXiv},
      primaryClass={cs.AI},
      url={https://arxiv.org/abs/2504.00424}, 
}

@misc{modelmerging,
      title={Unlocking Efficient Long-to-Short LLM Reasoning with Model Merging}, 
      author={Han Wu and Yuxuan Yao and Shuqi Liu and Zehua Liu and Xiaojin Fu and Xiongwei Han and Xing Li and Hui-Ling Zhen and Tao Zhong and Mingxuan Yuan},
      year={2025},
      eprint={2503.20641},
      archivePrefix={arXiv},
      primaryClass={cs.CL},
      url={https://arxiv.org/abs/2503.20641}, 
}

@misc{Claude3.7,
    title = {Anthropic. Claude 3.7 sonnet},
    author = {Anthropic},
    year = {2025},
    url = {https://www.anthropic.com/news/claude-3-7-sonnet},
    note = {Accessed on March 10, 2025}
}

@misc{Sketch-of-Thought,
      title={Sketch-of-Thought: Efficient LLM Reasoning with Adaptive Cognitive-Inspired Sketching}, 
      author={Simon A. Aytes and Jinheon Baek and Sung Ju Hwang},
      year={2025},
      eprint={2503.05179},
      archivePrefix={arXiv},
      primaryClass={cs.CL},
      url={https://arxiv.org/abs/2503.05179}, 
}

@misc{confidence,
      title={Learning to Route LLMs with Confidence Tokens}, 
      author={Yu-Neng Chuang and Helen Zhou and Prathusha Kameswara Sarma and Parikshit Gopalan and John Boccio and Sara Bolouki and Xia Hu},
      year={2025},
      eprint={2410.13284},
      archivePrefix={arXiv},
      primaryClass={cs.CL},
      url={https://arxiv.org/abs/2410.13284}, 
}

@misc{RouteLLMs,
      title={RouteLLM: Learning to Route LLMs with Preference Data}, 
      author={Isaac Ong and Amjad Almahairi and Vincent Wu and Wei-Lin Chiang and Tianhao Wu and Joseph E. Gonzalez and M Waleed Kadous and Ion Stoica},
      year={2025},
      eprint={2406.18665},
      archivePrefix={arXiv},
      primaryClass={cs.LG},
      url={https://arxiv.org/abs/2406.18665}, 
}

@misc{THOUGHTTERMINATOR,
      title={THOUGHTTERMINATOR: Benchmarking, Calibrating, and Mitigating Overthinking in Reasoning Models}, 
      author={Xiao Pu and Michael Saxon and Wenyue Hua and William Yang Wang},
      year={2025},
      eprint={2504.13367},
      archivePrefix={arXiv},
      primaryClass={cs.CL},
      url={https://arxiv.org/abs/2504.13367}, 
}

@misc{L1,
      title={L1: Controlling How Long A Reasoning Model Thinks With Reinforcement Learning}, 
      author={Pranjal Aggarwal and Sean Welleck},
      year={2025},
      eprint={2503.04697},
      archivePrefix={arXiv},
      primaryClass={cs.CL},
      url={https://arxiv.org/abs/2503.04697}, 
}

@misc{longsurvey,
      title={Towards Reasoning Era: A Survey of Long Chain-of-Thought for Reasoning Large Language Models}, 
      author={Qiguang Chen and Libo Qin and Jinhao Liu and Dengyun Peng and Jiannan Guan and Peng Wang and Mengkang Hu and Yuhang Zhou and Te Gao and Wanxiang Che},
      year={2025},
      eprint={2503.09567},
      archivePrefix={arXiv},
      primaryClass={cs.AI},
      url={https://arxiv.org/abs/2503.09567}, 
}

@misc{AIME2024,
    title = {In American Invitational
 Mathematics Examination},
    author = {MAA. American invitational mathematics examination},
    year = {2024},
    url ={https://maa.org/math-competitions/american-invitational-mathematics-examination-aime},
    note = {2024, February 2024}
}

@article{math500,
      title={Let's Verify Step by Step}, 
      author={Lightman, Hunter and Kosaraju, Vineet and Burda, Yura and Edwards, Harri and Baker, Bowen and Lee, Teddy and Leike, Jan and Schulman, John and Sutskever, Ilya and Cobbe, Karl},
      journal={arXiv preprint arXiv:2305.20050},
      year={2023}
}

@inproceedings{math,
 author = {Hendrycks, Dan and Burns, Collin and Kadavath, Saurav and Arora, Akul and Basart, Steven and Tang, Eric and Song, Dawn and Steinhardt, Jacob},
 booktitle = {Proceedings of the Neural Information Processing Systems Track on Datasets and Benchmarks},
 editor = {J. Vanschoren and S. Yeung},
 pages = {},
 title = {Measuring Mathematical Problem Solving With the MATH Dataset},
 url = {https://datasets-benchmarks-proceedings.neurips.cc/paper_files/paper/2021/file/be83ab3ecd0db773eb2dc1b0a17836a1-Paper-round2.pdf},
 volume = {1},
 year = {2021}
}

@misc{AMC23,
    title = {AIMO Validation AMC},
    author = {AI-MO},
    year = {2025},
    url ={https://huggingface.co/datasets/AI-MO/aimo-validation-amc},
    note = {2025, February 2025}
}

@inproceedings{OlympiadBench,
  author       = {Chaoqun He and
                  Renjie Luo and
                  Yuzhuo Bai and
                  Shengding Hu and
                  Zhen Leng Thai and
                  Junhao Shen and
                  Jinyi Hu and
                  Xu Han and
                  Yujie Huang and
                  Yuxiang Zhang and
                  Jie Liu and
                  Lei Qi and
                  Zhiyuan Liu and
                  Maosong Sun},
  editor       = {Lun{-}Wei Ku and
                  Andre Martins and
                  Vivek Srikumar},
  title        = {OlympiadBench: {A} Challenging Benchmark for Promoting {AGI} with
                  Olympiad-Level Bilingual Multimodal Scientific Problems},
  booktitle    = {Proceedings of the 62nd Annual Meeting of the Association for Computational
                  Linguistics (Volume 1: Long Papers), {ACL} 2024, Bangkok, Thailand,
                  August 11-16, 2024},
  pages        = {3828--3850},
  publisher    = {Association for Computational Linguistics},
  year         = {2024},
  url          = {https://doi.org/10.18653/v1/2024.acl-long.211},
  doi          = {10.18653/V1/2024.ACL-LONG.211},
  bibsource    = {dblp computer science bibliography, https://dblp.org}
}

@article{verl,
  title   = {HybridFlow: A Flexible and Efficient RLHF Framework},
  author  = {Guangming Sheng and Chi Zhang and Zilingfeng Ye and Xibin Wu and Wang Zhang and Ru Zhang and Yanghua Peng and Haibin Lin and Chuan Wu},
  year    = {2024},
  journal = {arXiv preprint arXiv: 2409.19256}
}

@misc{ThinkPrune,
      title={ThinkPrune: Pruning Long Chain-of-Thought of LLMs via Reinforcement Learning}, 
      author={Bairu Hou and Yang Zhang and Jiabao Ji and Yujian Liu and Kaizhi Qian and Jacob Andreas and Shiyu Chang},
      year={2025},
      eprint={2504.01296},
      archivePrefix={arXiv},
      primaryClass={cs.CL},
      url={https://arxiv.org/abs/2504.01296}, 
}

@misc{yang2025dynamicearlyexitreasoning,
      title={Dynamic Early Exit in Reasoning Models}, 
      author={Chenxu Yang and Qingyi Si and Yongjie Duan and Zheliang Zhu and Chenyu Zhu and Qiaowei Li and Zheng Lin and Li Cao and Weiping Wang},
      year={2025},
      eprint={2504.15895},
      archivePrefix={arXiv},
      primaryClass={cs.CL},
      url={https://arxiv.org/abs/2504.15895}, 
}
